 \documentclass[pmlr,twocolumn,10pt]{jmlr} 

\usepackage{siunitx}

\newcommand{\equal}[1]{{\hypersetup{linkcolor=black}\thanks{#1}}}

\theorembodyfont{\upshape}
\theoremheaderfont{\scshape}
\theorempostheader{:}
\theoremsep{\newline}

\jmlrvolume{297}
\jmlryear{2025}
\jmlrworkshop{Machine Learning for Health (ML4H) 2025} 

 \title[STAMP: Spatial-Temporal Adapter with Multi-Head Pooling]{STAMP: Spatial-Temporal Adapter with Multi-Head Pooling}

\author{%
\Name{Brad Shook}\equal{Corresponding author.}\footnotemark[2] \Email{bshook@andrew.cmu.edu}\\
\Name{Abby Turner}\footnotemark[2] \Email{abigailt@andrew.cmu.edu}\\
\Name{Jieshi Chen}\footnotemark[2] \Email{jieshic@andrew.cmu.edu}\\
\Name{Michał Wiliński}\footnotemark[2] \Email{mwilinsk@andrew.cmu.edu}\\
\Name{Mononito Goswami}\footnotemark[2] \Email{mgoswami@andrew.cmu.edu}\\
\Name{Jonathan Elmer}\footnotemark[3] \Email{elmerjp@upmc.edu}\\
\Name{Artur Dubrawski}\footnotemark[2] \Email{awd@cs.cmu.edu}\\
\addr \textsuperscript{†}Carnegie Mellon University, Pittsburgh, PA, USA\\
\addr \textsuperscript{‡}University of Pittsburgh School of Medicine, Pittsburgh, PA, USA
}

\begin{document}

\maketitle

\begin{abstract}
Time series foundation models (TSFMs) pretrained on data from multiple domains have shown strong performance on diverse modeling tasks. Various efforts have been made to develop foundation models specific to electroencephalography (EEG) data, which records brain electrical activity as time series. However, no comparative analysis of EEG-specific foundation models (EEGFMs) versus general TSFMs has been performed on EEG-specific tasks. We introduce a novel \textbf{S}patial-\textbf{T}emporal \textbf{A}dapter with \textbf{M}ulti-Head \textbf{P}ooling (\textbf{STAMP}), which leverages univariate embeddings produced by a general TSFM, implicitly models spatial-temporal characteristics of EEG data, and achieves performance comparable to state-of-the-art EEGFMs. A comprehensive analysis is performed on 8 benchmark datasets of clinical tasks using EEG for classification, along with ablation studies. Our proposed adapter is lightweight in trainable parameters and flexible in the inputs it can accommodate, supporting easy modeling of EEG data using TSFMs.

\end{abstract}

\paragraph*{Data and Code Availability}

We use 8 EEG benchmarking datasets, which are publicly available, to evaluate our method. We use four datasets for ablation experiments: SHU-MI \citep{ma_shu_2022}, MentalArithmetic \citep{goldberger_physiobank_2000, zyma_stress_2019}, BCIC-IV-2a \citep{brunner_bci2a_nodate}, PhysioNet-MI \citep{goldberger_physiobank_2000, schalk_bci2000_2004}. For final evaluation of our methods, we use those same datasets, in addition to Mumtaz2016 \citep{mumtaz_mdd_2016}, SEED-V \citep{liu_seedv_2022}, TUEV \citep{obeid_tuh_2016}, and FACED \citep{chen_faced_2023}. Details about these datasets are provided in Appendix \ref{apd:datasets}. Our code is publicly available at \url{https://github.com/autonlab/STAMP}. 

\paragraph*{Institutional Review Board (IRB)} This work does not require IRB approval since we performed experiments on publicly available datasets. 

\section{Introduction}
\label{sec:intro}

Foundation models can achieve strong performance on diverse modeling tasks by leveraging large-scale pretraining using data from a particular modality.
Due to an abundance of modeling applications and available data, one modality of particular interest is time series. Various efforts have been made to build general-purpose time series foundation models (TSFMs) \citep{goswami_moment_2024, ansari_chronos_2024, das_decoder-only_2024}, pretrained on data from multiple domains. Concurrently, there have been efforts to build foundation models specifically for electroencephalography (EEG) data, where electrical activity of the brain is measured and recorded as time series~\citep{wang_cbramod_2025, wang_eegpt_2024, jiang_large_2024}. To our knowledge, no comparative analysis of EEG-specific foundation models (EEGFMs) versus TSFMs has yet been performed on EEG-specific tasks. 

EEG measures electrical activity generated by neurons in the brain's cortex by placing electrodes on the scalp in standard locations. They are connected to an amplifier that applies basic bandpass filtering and converted to digital signals for recording. The resulting signals are spatiotemporal in nature and provide valuable information on brain health and activity, useful for predicting emotion, sleep stage, seizure activity, and Alzheimer's disease \citep{craik_deep_2019}. 

While most EEGFMs attempt to model both spatial and temporal dependencies, TSFMs typically model univariate time series
and are not naturally effective at EEG-related tasks. We introduce \textbf{STAMP} (\textbf{S}patial-\textbf{T}emporal \textbf{A}dapter with \textbf{M}ulti-Head \textbf{P}ooling), a lightweight, flexible adapter for use on top of general-purpose TSFMs that achieves performance comparable to state-of-the-art EEGFMs across multiple EEG classification tasks. The adapter enables direct use of existing pre-trained TSFMs, potentially reducing solution development costs and time, and greatly outperforms naive mean pooling with TSFMs, which yields near-random performance.
STAMP leverages univariate embeddings produced by a TSFM and implicitly models relationships across both spatial and temporal dimensions present in EEG. There are three main components within the adapter: 1) positional encodings (PEs) that enable our model to earmark spatial and temporal locations associated to TSFM embeddings, 2) a criss-cross gated MLP (CC-GMLP) that captures spatial and temporal relationships between the resulting embeddings, and 3) multi-head attention pooling (MHAP; \cite{india_self_2019, zhao_multi-query_2022}) that extracts relevant information to produce a final prediction. While EEGFMs have millions, and sometimes tens of millions, of trainable parameters, our adapter has a fraction of that (approximately 750 thousand), reducing data requirements. Since we freeze the parameters of the TSFM, embeddings can be generated once for a dataset and the hyperparameters of our adapter can be easily tuned. We perform extensive ablation experiments to compare various choices for our adapter and justify the need for each component. Next, we run a full evaluation of STAMP and compare it against 2 EEGFMs and 2 non-foundation EEG models. Lastly, we compare STAMP when used with embeddings from different TSFMs. 

\begin{table*}[htbp]
\floatconts
  {table:fm_comparison}
  {\caption{Comparison between general-purpose TSFMs and EEGFMs.}}%
  {%
    \begin{tabular}{|>{\centering\arraybackslash}p{3cm}|p{6cm}|p{6cm}|}
    \hline
    \abovestrut{2.2ex}\bfseries Aspect & \bfseries TSFM & \bfseries EEGFM \\\hline
    Pretraining Data & Diverse datasets from multiple domains (energy, weather, finance, health, etc.) & EEG datasets from various clinical domains
    \\\hline
    Data Characteristics & Varying sampling rates, seasonality, diverse patterns & High frequency sampling, clinically significant artifacts, spatial relationships
    \\\hline
    Modeling Architectures & General architectures which are domain-agnostic such as transformers & Architectures that leverage the characteristics of EEG (Fourier spectrum prediction, criss-cross transformers, etc.)
    \\\hline
    Typical Tasks & Forecasting, anomaly detection, imputation, classification, and regression & Classification and regression
    \\\hline
    \end{tabular}
  }
\end{table*}

\section{Related Works}
\label{sec:related_works}

\subsection{Time Series Foundation Models}
TSFMs gained traction by demonstrating that large-scale pretraining in this modality unlocks abilities that transfer across various time series tasks and diverse domains \citep{rasul_lag-llama_2023, das_decoder-only_2024, ansari_chronos_2024, goswami_moment_2024}. Their pretraining data includes millions of time series and spans multiple domains such as finance, healthcare, energy, weather, and more. Most of the focus on TSFMs has been on their architectures and pretraining using self-supervised learning (SSL), while less attention has been paid to adapting and fine-tuning them for downstream tasks. For example, \cite{goswami_moment_2024} pretrain their model using SSL and a reconstruction head. Then, to perform a downstream task of classification, a single linear layer replaces the reconstruction head and is further fine-tuned on a per-dataset basis.

\subsection{EEG Foundation Models}
At the same time that TSFMs were being developed, EEGFMs were also being built. EEGFMs have similar goals to TSFMs, however,  
their pretraining and use have been restricted to only EEG data and tasks. Early EEGFMs, such as Neuro-GPT \citep{cui_neuro-gpt_2024}, followed similar architectures as TSFMs and did not leverage any specific characteristics present in EEG data. 

Recent EEGFMs attempt to address this. For example, LaBraM \citep{jiang_large_2024} employs multiple EEG-specific components such as spatial and temporal embeddings and Fourier spectrum prediction. CBraMod \citep{wang_cbramod_2025} uses a different approach, in particular, asymmetric conditional positional encoding (ACPE) and a criss-cross transformer (CC-TF). ACPE encodes spatial and temporal positional information while prioritizing short-range temporal information and long-range spatial information. The CC-TF applies self-attention spatially and temporally separately, rather than across a single axis. 

Another difference between TSFMs and EEGFMs is the sampling rate of the pretraining data. Specifically, \cite{goswami_moment_2024}'s pretraining data contains datasets with varying sampling rates ranging from 15 minutes to daily to weekly, with only a few high frequency datasets. In contrast, EEGFMs often resample their pretraining data to a single sampling rate such as 200 Hz for \cite{jiang_large_2024} and \cite{wang_cbramod_2025}. Table \ref{table:fm_comparison} shows a comprehensive comparison between the two types of foundation models.

Despite the related architectures and goals of TSFMs and EEGFMs, to our knowledge, no research has been published comparing the two on EEG tasks. 

\subsection{Token Mixing and Aggregation}

As will be further detailed in our methods section, adapting a TSFM to handle EEG data requires strategies for token ``mixing" and ``aggregation." A token, for our purposes, refers to a discrete segment of time series processed by the TSFM. Since EEG data are available from different spatial channels and over potentially long periods of time, EEG data is decomposed into a ``spatialtemporal grid" of tokens. Using this grid for predictions requires both ``mixing" (capturing relationships between tokens throughout the grid) and ``aggregation" (summarizing across the grid for a final prediction).

Transformer encoders 
\citep{vaswani_attention_2017} are an obvious choice for mixing tokens due to their success in language modeling. The problem with a standard transformer encoder is that attention is applied across all tokens, rather than taking advantage of the distinct spatial and temporal patterns of EEG. An option aiming to better leverage EEG structure is the CC-TF from \cite{wang_cbramod_2025}. The authors demonstrated that for EEG tasks, the CC-TF outperforms the standard transformer encoder architecture. An architecturally simpler option for token mixing is a gated MLP (GMLP) \citep{liu_pay_2021}. GMLPs have shown competitive performance to transformers in both language and image modeling. The standard GMLP formulation models interactions across all tokens. 

For token aggregation, naive strategies include averaging tokens before input to a prediction head, or averaging predictions made on each token. These strategies have the disadvantage of treating each token as carrying information of equal value: A more advanced solution designed to extract and weight features from each token is multi-head attention pooling, previously introduced by \cite{india_self_2019} and expanded upon by \cite{zhao_multi-query_2022}.

\section{Methods}
\label{sec:methods}

\begin{figure*}[htbp]
\floatconts
  {fig:main_diagram}
  {\caption{A diagram showing how EEG data is processed by MOMENT and STAMP. The EEG data is separated into tokens, which are embedded using MOMENT before positional encoding is applied. The resulting tokens are passed through the CC-GMLP, where spatial and temporal relationships are incorporated into embeddings. MHAP then determines relevant features and generates final predictions by projecting embeddings into lower dimensional spaces.}}
  {\includegraphics[width=\linewidth]{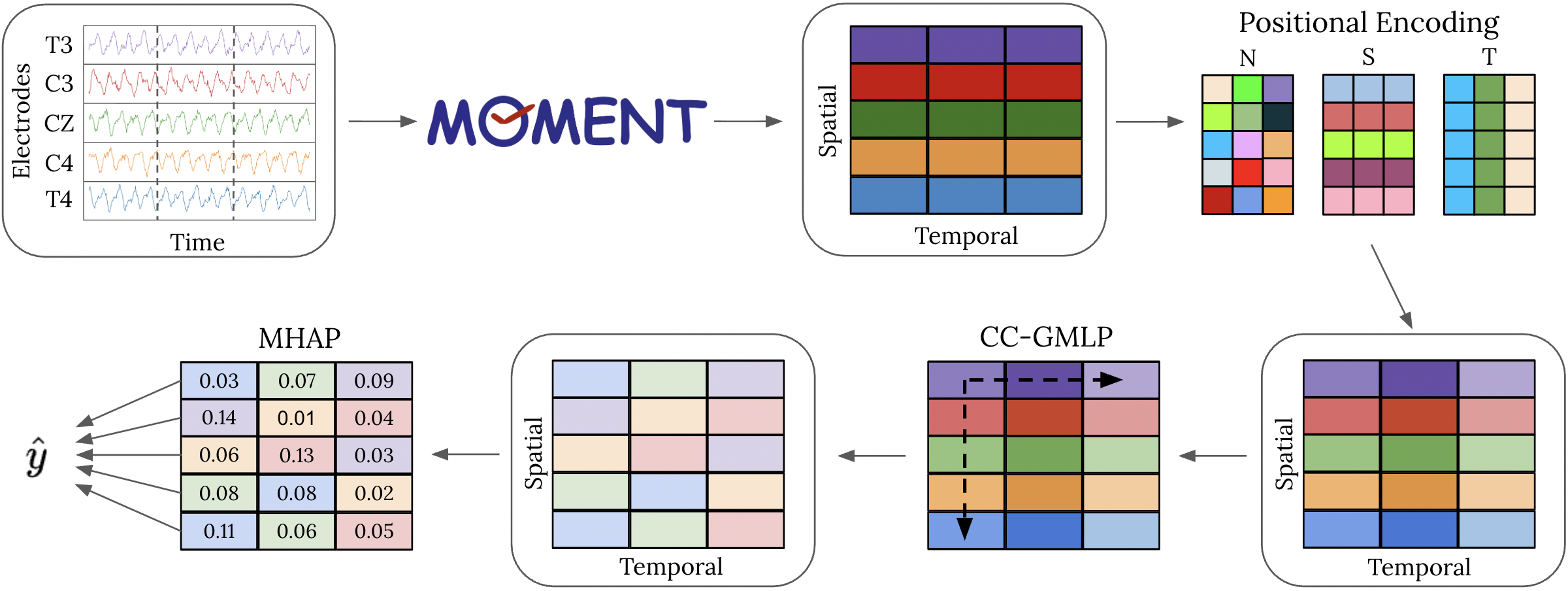}}
\end{figure*}

\subsection{Embedding Generation}
Let $X \in \mathbb{R}^{S \times \mathcal{T}}$ represent a collection of time series with $S$ spatial channels containing $\mathcal{T}$ observations each. For each channel, we partition the associated time series of length $\mathcal{T}=TK$ into sequences of length~$K=200$ (corresponding to 1 second of data), yielding $X\in \mathbb{R}^{S \times T \times K}$. Each sequence or ``token" serves as an input vector to a TSFM with frozen parameters. The output is an embedding of each token in $\mathbb{R}^{\ell}$, yielding a grid of TSFM embeddings $E \in \mathbb{R}^{S \times T \times \ell}$.
Unless otherwise noted, MOMENT Large \citep{goswami_moment_2024} is used as the TSFM, yielding embeddings in $\mathbb{R}^{\ell}$ for $\ell=1024$. 

To reduce the total number of parameters in our adapter, we start by applying a learnable linear 
mapping $W$ from $\mathbb{R}^{\ell}$ to $\mathbb{R}^D$ to each embedding. Using $\cdot$ to denote the application of such a mapping along the appropriate dimension when unambiguous, this yields a grid $E' = E \cdot W \in \mathbb{R}^{S \times T \times D}$ of embeddings with reduced dimensionality.

\subsection{Positional Encoding}
In order to encode spatial, temporal, and token positional information, we next add learnable positional embeddings to the embeddings in our grid $E'$. 

In particular, we learn token-wise positional embeddings $p_{ij}\in\mathbb{R}^D$ for $(i,j)\in S\times T$, unique to each embedding in our grid, along with spatial-wise and temporal-wise positional embeddings $s_i, t_j \in \mathbb{R}^D$ for $i\in S, j\in T$, unique to an embedding's spatial and temporal position in our grid, respectively. Thus, if $e'_{ij} \in \mathbb{R}^D$ denotes one embedding in our grid $E'\in \mathbb{R}^{S\times T \times D}$, from this stage we obtain a modified grid $\tilde{E} \in \mathbb{R}^{S\times T \times D}$ with entries $\tilde{e}_{ij} = e'_{ij} + p_{ij}+s_i+t_j$. A visualization of this procedure is seen in \figureref{fig:main_diagram}.

While it may seem more natural to include only the $|S\times T|$ token embeddings (i.e. set $\tilde{e}_{ij} = e'_{ij} + p_{ij}$) or only the $|S|+|T|$ spatial and temporal embeddings (i.e. set $\tilde{e}_{ij} = e'_{ij} + s_i+t_j$), in our ablation studies, we find that the structure enforced by using all three typically improves performance. 
  
\subsection{Criss-Cross GMLP}

In order to model spatial and temporal relationships, we use a novel criss-cross GMLP (CC-GMLP) inspired by the original GMLP \citep{liu_pay_2021} and the CC-TF \citep{wang_cbramod_2025}. We found the criss-cross architecture used in CBraMod appealing due to its separate modeling of spatial and temporal relationships, which we hoped would enable parameter-efficient learning of cross-channel relationships independent of a particular signal’s temporal evolution. In line with this intuition, during the development of STAMP, we found that the criss-cross architecture often improved performance for both transformers and GMLP, as highlighted by \figureref{fig:token_mixing}. The CC-GMLP is made up of $L$ blocks, each block taking inputs $\tilde{E} \in \mathbb{R}^{S\times T \times D}$ and computing:

\begin{align}
    Z = \sigma(\tilde{E}\cdot U)\\
    \tilde{Z}_T=g_T(Z),\ \tilde{Z}_S=g_S(Z)\\
    \tilde{Z} = \text{Concat}(\tilde{Z}_T, \tilde{Z}_S)\\
    \hat{E} =\tilde{Z} \cdot V
\end{align}

where $\sigma$ is the GELU activation function, $g_T(\cdot)$ and $g_S(\cdot)$ are the temporal gating unit and spatial gating unit, respectively, and $U : \mathbb{R}^D \rightarrow \mathbb{R}^h$, $V : \mathbb{R}^h \rightarrow \mathbb{R}^D$ are linear maps applied along the token dimension. 
The temporal and spatial gating units follow the same formulation as the original GMLP, Equation \ref{basic_gmlp_sgu}, except that the linear projection is applied only along the respective axis. For example, the spatial gating unit separates $Z \in \mathbb{R}^{S\times T \times h}$ into $Z_1,Z_2 \in \mathbb{R}^{S \times T \times \frac{h}{2}}$ and, using $\odot$ to denote element-wise multiplication, learns a linear mapping $W : \mathbb{R}^{S} \rightarrow \mathbb{R}^{S}$ along the spatial dimension so that 
\begin{align}
\label{basic_gmlp_sgu}
    g_S(Z) = Z_1 \odot (W \cdot Z_2).
\end{align}

In accordance with the original GMLP implementation, we initialize $W$ with mean 0 and standard deviation $10^{-6}$.
Note that layer normalization is applied to the input of each block and a residual connection is used, such that the final block output is $(\hat{E}+\tilde{E})\in \mathbb{R}^{S\times T\times D}$.

The GMLP paper \citep{liu_pay_2021} reports that their model does not require positional embeddings because that information is captured by $g(\cdot)$. However, we have found that positional embeddings significantly improve the performance of our adapter, 
as demonstrated by \figureref{fig:pe}. We hypothesize that positional encoding is specifically helpful for the task of EEG modeling due to the extensive spatial and temporal correlations within the data. 

\subsection{Multi-Head Attention Pooling}
Our adapter uses a variant of multi-head attention pooling (MHAP) to aggregate the tokens \citep{zhao_multi-query_2022}.
MHAP allows the adapter to learn to weight the importance of each token with respect to the final prediction. Given our grid of mixed token embeddings $\hat{E} \in \mathbb{R}^{S\times T\times D}$, $A$ heads, $Q$ queries per head, and $d = D / A$, MHAP begins by linearly mapping tokens to lower-dimensional spaces via $W_a : \mathbb{R}^D \rightarrow \mathbb{R}^d$ for each $a\in A$. Subsequently, the projected tokens for a given $a$ are measured against queries $r_{a,q} \in \mathbb{R}^{d
\times 1}$ for each $q\in Q$. 
The resulting scores determine an aggregate vector for each $q$ via weighted summation of token projections over the spatiotemporal grid. The sum of these scores are further used to measure the query's overall importance in a final aggregation step across queries. Formally, for each $a\in A$ we obtain a vector $z_a$ aggregated over all queries as follows:
\vspace{-1ex}
\begin{align}
    &\textbf{Project)} \ H_a = \left(\hat{E}\cdot W_a\right) \in \mathbb{R}^{S\times T \times d} \\
    &\textbf{For each $q\in Q$:} \\
    &\hspace{2ex} \textbf{     Attend)} \ \alpha_{a,q} = \text{softmax}(\frac{H_a \cdot r_{a,q}}{\sqrt{d}}) \in \mathbb{R}^{S\times T} \\
   &\hspace{2ex} \textbf{     Pool)} \ u_{a,q} = (\sum_{i,j \in S \times T} \alpha_{a,q}^{(i,j)} H_a^{(i,j)} ) \in \mathbb{R}^d \\
       &\hspace{2ex} \textbf{     Weight)} \ \beta_{a,q} = (\sum_{i,j \in S\times T}\alpha_{a,q}^{(i,j)})\in \mathbb{R} \\
    &\textbf{Combine) } \ \beta_a=\text{softmax}\!\big((\beta_{a,q})_{q=1}^{Q}\big)\in\mathbb{R}^Q  \\
      &\hspace{12ex} \ z_a = \sum^Q_{q=1} \beta_{a,q} u_{a,q} \in \mathbb{R}^d.
\end{align}

The final vector in $\mathbb{R}^d$ for each head is concatenated into a summary in $[z_1, \cdots, z_a]=z\in\mathbb{R}^D$. Adding a residual connection to the mean mixed token embedding $\hat{e} = \left( \frac{1}{|S\times T|} \sum_{i,j \in S\times T} \hat{E}^{(i,j)} \right) \in \mathbb{R}^D$, final predictions are produced with a linear map $W: \mathbb{R}^D \rightarrow \mathbb{R}^n$ via $\hat{y}=\text{softmax}(W \cdot (\lambda z+(1-\lambda) \hat{e})).$

\section{Experimental Setup}
\label{sec:experiments}

\subsection{Evaluation Datasets}
We evaluate our approach on 8 common EEG datasets. The datasets and their characteristics are reported in \tableref{tab:datasets}. In order to make our evaluations comparable to other EEGFMs, our datasets are preprocessed using the same procedure as CBraMod. In short summary, the EEG signals in each dataset are noise filtered, resampled to 200 Hz, and split into training, validation, and test splits. For more details on pre-processing, see \cite{wang_cbramod_2025}. 

\subsection{Baseline Methods}
Multiple baseline methods are used for comparison. Specifically, we compare against two non-foundation EEG models: EEG Conformer \citep{eeg_conformer} and ST-Transformer \citep{st_transformer}. As for EEGFMs, we include CBraMod \citep{wang_cbramod_2025} and LaBraM-Base \citep{jiang_large_2024}. During our experiments, we found that running CBraMod with the suggested default configurations sometimes yielded performance metrics lower than those reported in their paper\footnote{Multiple users have reported similar issues reproducing CBraMod results on certain datasets: \url{https://github.com/wjq-learning/CBraMod/issues}}. Given this discrepancy, we report only the reproduced CBraMod performance. Due to computational limitations, we do not reproduce the results of the other baselines, but rather report the performance metrics and parameter counts detailed previously by~\cite{wang_cbramod_2025}, which were obtained as described in Section 3.2 of that work.
\subsection{Evaluation Metrics}
Following the work of \cite{jiang_large_2024} and \cite{wang_cbramod_2025}, we report the same evaluation metrics. For binary classification, we use Balanced Accuracy, AUC-PR, and AUROC as evaluation metrics, with AUROC as our monitor metric during validation. Note that a threshold of 0.5 is used to calculate the balanced accuracy in the binary setting, consistent with \cite{wang_cbramod_2025}. For multiclass classification, we use Balanced Accuracy, Cohen's Kappa Score, and Weighted F1, with Cohen's Kappa Score as our monitor metric for validation.

\subsection{Experiments}
All of our experiments were run using Pytorch 2.6.0 and CUDA 12.4. Embeddings were generated in parallel using multiple NVIDIA GeForce RTX 2080 Ti GPUs each with 12GB of VRAM. Most experiments were run on a single NVIDIA GeForce RTX 2080 Ti GPU (with 12GB of VRAM), but a minority of the experiments used an NVIDIA RTX A6000 GPU with 48GB of VRAM or an NVIDIA Tesla V100 with 32GB of VRAM. 

In order to justify the component choices in our adapter, we performed ablation studies that examined the impact of the positional encoding (PE), token mixing, and token aggregation components. Each ablation study fixes two of our three components, namely (token, spatial, and temporal) PE, CC-GMLP, and MHAP, and varies one component of interest. The parameter counts for different variants are averaged across the datasets and noted in the corresponding figure caption. The frozen parameters of MOMENT are not included in our parameter counts. We considered tuning a subset of MOMENT's parameters alongside the adapter, for example using popular low rank adaptation (LoRA) introduced by \cite{hu2022lora}, but did not observe performance improvements meriting the additional complexity (see \figureref{fig:finetune}).

In the positional encoding ablation study, we perform experiments with various PE options. These options are no PE, token PE (N), spatial and temporal PE (ST), or token, spatial, and temporal PE (NST). The token mixing ablation experiment varies the token mixing component. We compare the basic GMLP (B-GMLP), CC-GMLP, basic transformer (B-TF), and CC-TF. The token aggregation ablation experiment compares mean pooling and MHAP. All ablations compare performance on the test split. For brevity, we mainly compare AUROC and Cohen's Kappa Score, however, detailed performance metrics are found in \tableref{tab:shu_phys,tab:mental_arithmetic_bcic,tab:tuev_mumtaz,tab:seedv_faced}. 

Each ablation experiment uses the same 3 randomly generated seeds. These seeds ensure that the experiments are reproducible and that all variability is consistent across experiments. The mean and standard deviation across the seeds is calculated for each performance metric and used as our reporting statistics. Details about our hyperparameters are discussed in Appendix \ref{apd:hyperparameters}. 

A full evaluation of STAMP is run using 5 randomly generated seeds (including the 3 used in the ablation studies). We made efforts to ensure that our evaluation methodology mirrored what is done by \cite{wang_cbramod_2025} so that the performance between different methods can be fairly compared. This includes ensuring that our training, validation, and test splits matched and that our performance evaluation was calculated the same. 

To demonstrate that the success of STAMP is not contingent on using MOMENT Large, we further evaluated performance when using embeddings from other TSFMs, specifically MOMENT Small and Base, Chronos Large \citep{ansari_chronos_2024}, and TSPulse \citep{ekambaram2025tspulsedualspacetiny}. These evaluations followed the same 5 seed regiment as previously described.

\section{Results and Analysis}
\label{sec:results}

\subsection{Positional Encoding Ablation}

\begin{figure}[htbp]
\floatconts
  {fig:pe}
  {\caption{Performance comparison between four positional encoding options: No PE (0.71M), PE-N (0.73M), PE-ST (0.72M), and PE-NST (0.74M). The value in parentheses indicates the average number of trainable parameters across the 4 datasets.}}
  {\includegraphics[width=\linewidth]{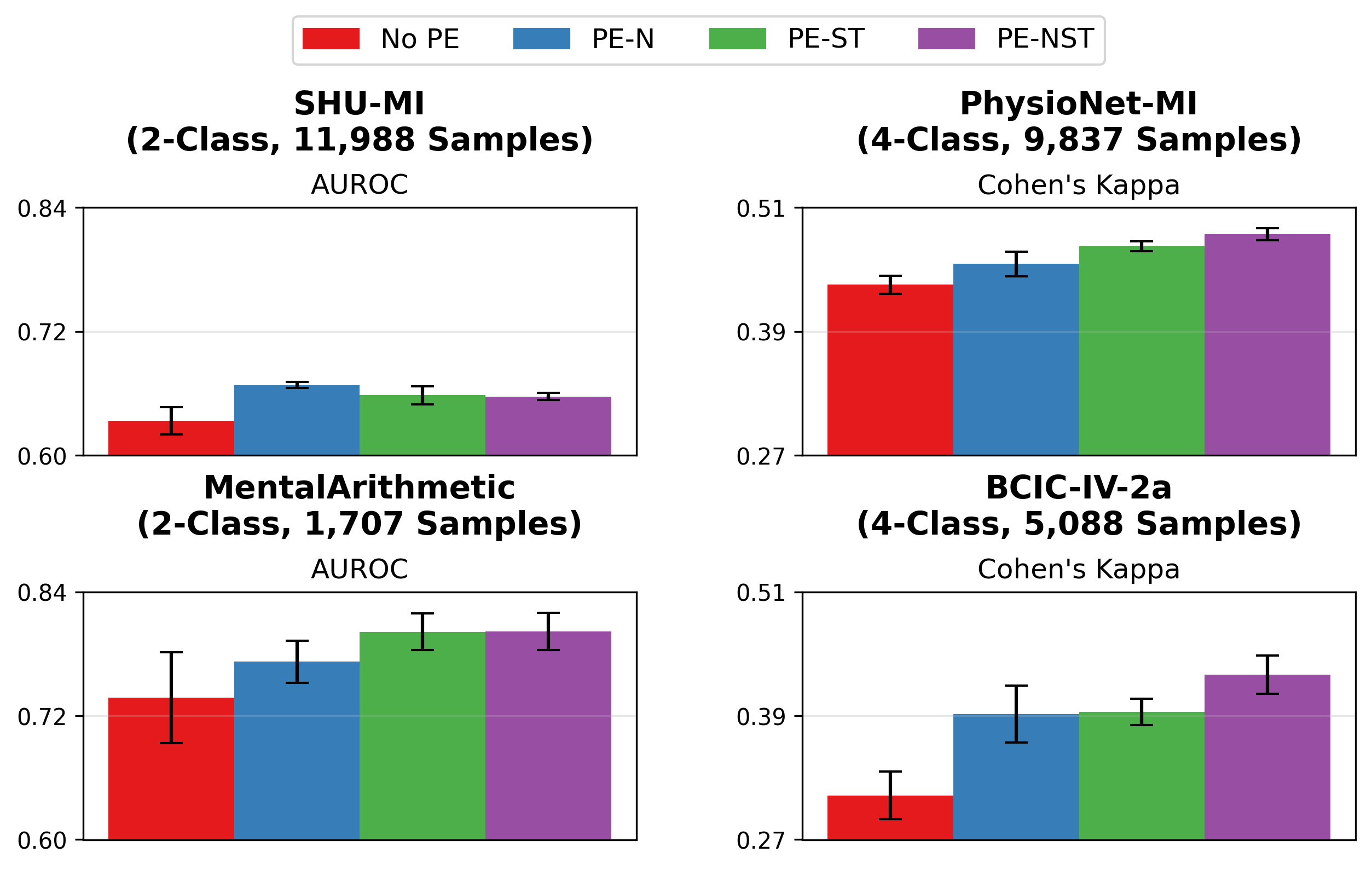}}
\end{figure}

Through our ablation of positional encoding (see \figureref{fig:pe}), we find that PE-NST yields the best performance in 3 of the 4 datasets. For the SHU-MI dataset, token-wise positional encoding outperformed the other options. In general, the more positional encoding used, the better the performance. The exclusion of positional encoding consistently results in the worst performance across each dataset, indicating that positional encoding is essential to the adapter. 

Our speculation is that positional encoding is particularly important for EEG data due to extensive spatiotemporal correlations and dependencies. PEs make both spatial and temporal locations associated with embeddings more explicit to the adapter. Competitive performance of PE-N and boosts in performance from PE-NST suggest that the adapter even benefits from the additional parameters and flexibility needed to encode token-wise (vs. only axis-wise) relationships.

\subsection{Token Mixing Ablation}
\begin{figure}[htbp]
\floatconts
  {fig:token_mixing}
  {\caption{Performance comparison between four different token mixer options: B-GMLP (0.79M), CC-GMLP (0.74M), B-TF (1.25M), and CC-TF (0.99M). The value in parentheses indicates the average number of trainable parameters across the 4 datasets.}}
  {\includegraphics[width=\linewidth]{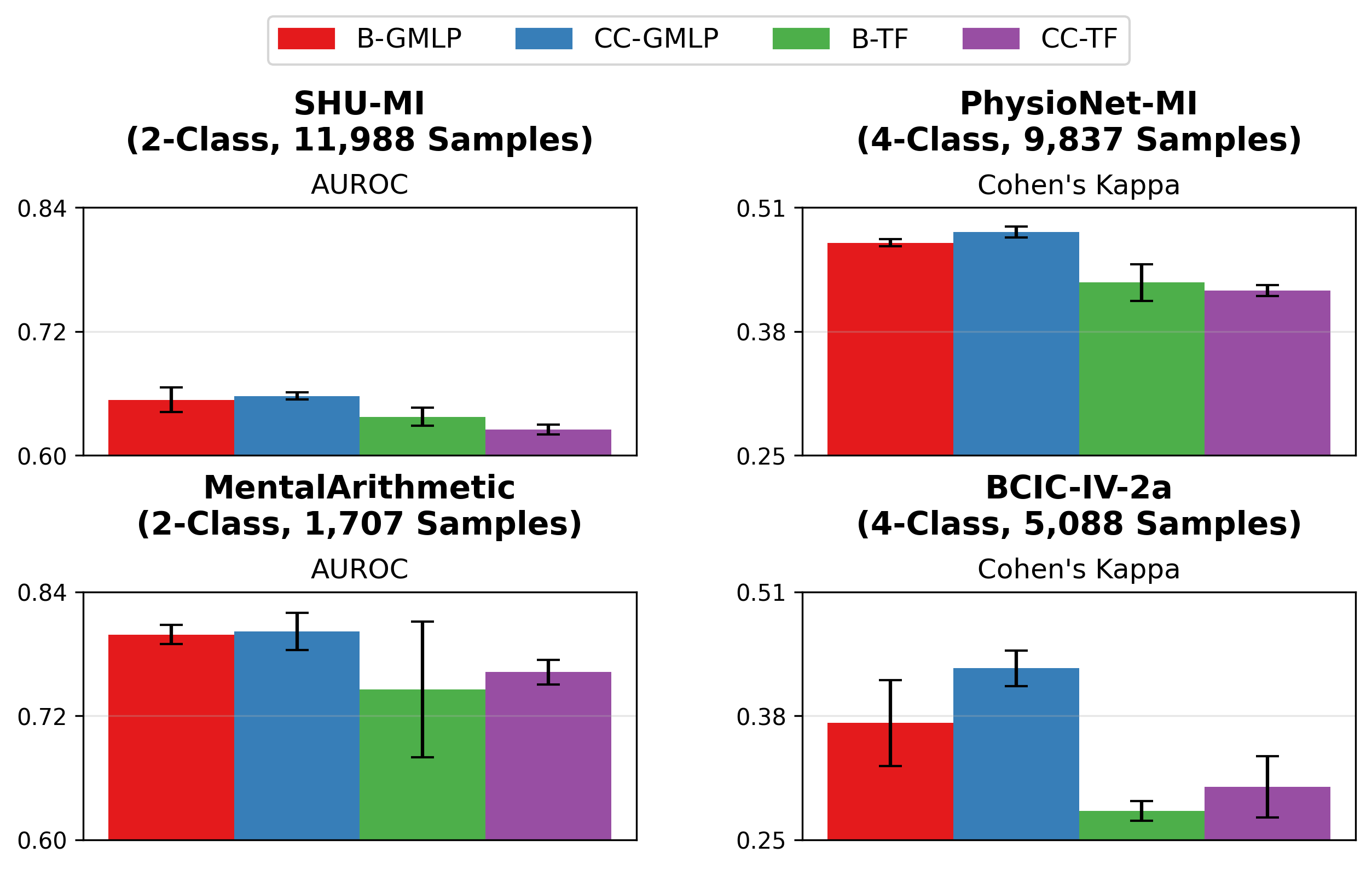}}
\end{figure}

In our ablation study comparing token mixing strategies, we see that CC-GMLP performs strongly across each dataset. Across all four datasets, the GMLP architecture performs better than its transformer counterpart.

\subsection{Token Aggregation Ablation}

\begin{figure}[htbp]
\floatconts
  {fig:token_agg}
  {\caption{Performance comparison between token aggregation strategies: mean pooling (0.70M) and MHAP (0.74M). The value in parentheses indicates the average number of trainable parameters across the 4 datasets.}}
  {\includegraphics[width=\linewidth]{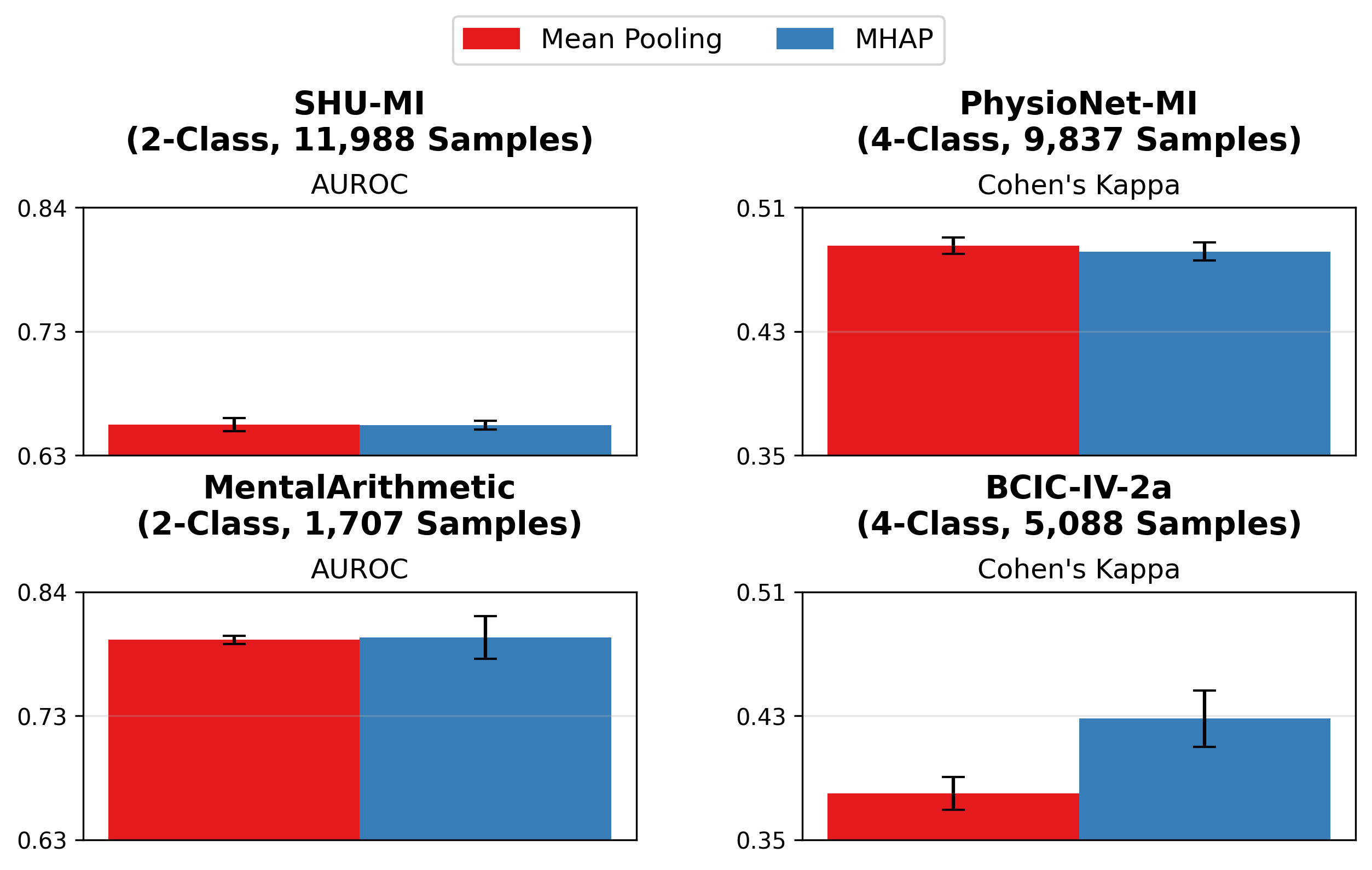}}
\end{figure}

Our last ablation study indicates that the choice of token aggregation strategy is less impactful than the positional encoding and token mixer choices. Specifically, performance between mean pooling and MHAP on every dataset except BCIC-IV-2a is similar. However, MHAP demonstrates a significant performance boost on the BCIC-IV-2a dataset, indicating that MHAP may be beneficial on other datasets. In \figureref{fig:moment_mean_pooling_vs_mhap_no_token_mixing}, we show that MHAP greatly outperforms mean pooling when a token mixing component is not used. We suspect that this performance difference is diminished when token mixing is included because MHAP and token mixing both implicitly model cross-token relationships. 
Since our CC-GMLP is able to capture cross-token relationships, mean pooling usually still performs competitively.

\subsection{Full Evaluation}
\begin{figure}[htbp]
\floatconts
  {fig:final_eval_2x2}
  {\caption{Performance comparison between the full evaluation of 5 methods: STAMP (0.74M), CBraMod (29M), LaBraM (5.8M), ST-Transformer (3.5M), and EEG Conformer (0.55M). The value in parentheses indicates the average number of trainable parameters across the 4 datasets.}}
  {\includegraphics[width=\linewidth]{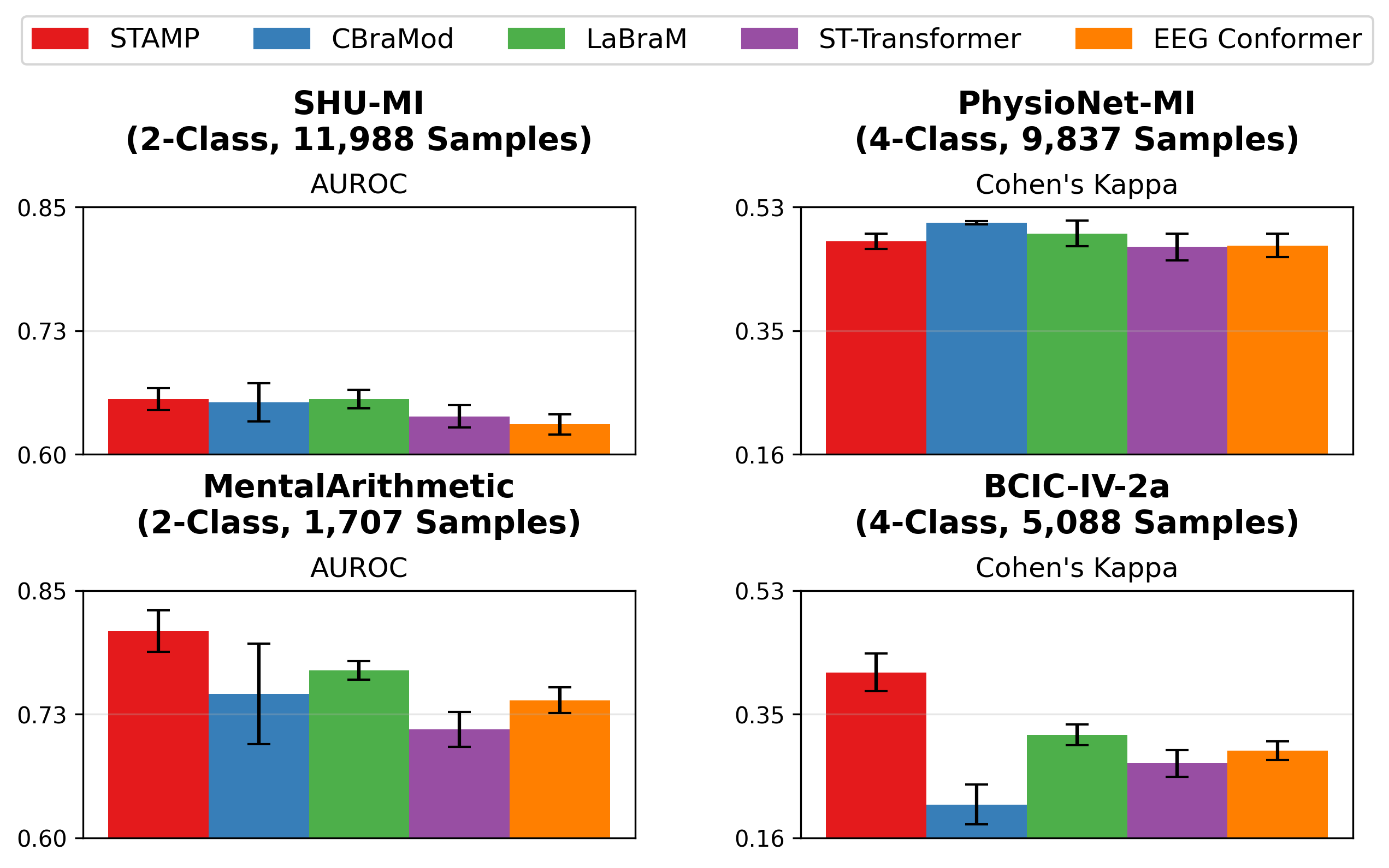}}
\end{figure}

\begin{table*}[htbp]
\centering
\caption{Performance comparison of different methods on SHU-MI and PhysioNet-MI datasets.}
\resizebox{\textwidth}{!}{%
\begin{tabular}{|l|r|ccc|ccc|}
\hline
 & & \multicolumn{3}{c|}{\textbf{SHU-MI (2-Class, 11,988 Samples)}} & \multicolumn{3}{c|}{\textbf{PhysioNet-MI (4-Class, 9,837 Samples)}} \\
\cline{3-8}
\textbf{Methods} & \textbf{\#Params} & \textbf{Balanced Acc.} & \textbf{AUC-PR} & \textbf{AUROC} & \textbf{Balanced Acc.} & \textbf{Cohen's Kappa} & \textbf{Weighted F1} \\
\hline
EEG Conformer & 0.55M & 0.5900 ± 0.0107 & 0.6370 ± 0.0093 & 0.6351 ± 0.0101 & 0.6049 ± 0.0104 & 0.4736 ± 0.0171 & 0.6062 ± 0.0095 \\

ST-Transformer & 3.5M & 0.5992 ± 0.0206 & 0.6394 ± 0.0122 & 0.6431 ± 0.0111 & 0.6035 ± 0.0081 & 0.4712 ± 0.0199 & 0.6053 ± 0.0075 \\
\hline
LaBraM & 5.8M & 0.6166 ± 0.0192 & 0.6761 ± 0.0083 & 0.6604 ± 0.0091 & 0.6173 ± 0.0122 & 0.4912 ± 0.0192 & 0.6177 ± 0.0141 \\

CBraMod & 25.5M/46M & 0.6043 ± 0.0069 & 0.6729 ± 0.0172 & 0.6572 ± 0.0191 & 0.6305 ± 0.0017 & 0.5072 ± 0.0023 & 0.6313 ± 0.0016 \\
\hline 
STAMP & 0.73M/0.78M & 0.5983 ± 0.0096 & 0.6630 ± 0.0128 & 0.6603 ± 0.0109 & 0.6098 ± 0.0084 & 0.4797 ± 0.0112 & 0.6111 ± 0.0102 \\
\hline
\end{tabular}
}
\label{tab:shu_phys}
\end{table*}

\begin{table*}[htbp]
\centering
\caption{Performance comparison of different methods on MentalArithmetic and BCIC-IV-2a datasets.}
\resizebox{\textwidth}{!}{%
\begin{tabular}{|l|r|ccc|ccc|}
\hline
 & & \multicolumn{3}{c|}{\textbf{MentalArithmetic (2-Class, 1,707 Samples)}} & \multicolumn{3}{c|}{\textbf{BCIC-IV-2a (4-Class, 5,088 Samples)}} \\
\cline{3-8}
\textbf{Methods} & \textbf{\#Params} & \textbf{Balanced Acc.} & \textbf{AUC-PR} & \textbf{AUROC} & \textbf{Balanced Acc.} & \textbf{Cohen's Kappa} & \textbf{Weighted F1} \\
\hline
EEG Conformer & 0.55M & 0.6805 ± 0.0123 & 0.5829 ± 0.0134 & 0.7424 ± 0.0128 & 0.4696 ± 0.0106 & 0.2924 ± 0.0141 & 0.4533 ± 0.0128 \\

ST-Transformer & 3.5M & 0.6631 ± 0.0173 & 0.5672 ± 0.0259 & 0.7132 ± 0.0174 & 0.4575 ± 0.0145 & 0.2733 ± 0.0198 & 0.4471 ± 0.0142 \\
\hline
LaBraM & 5.8M & 0.6909 ± 0.0125 & 0.5999 ± 0.0155 & 0.7721 ± 0.0093 & 0.4869 ± 0.0085 & 0.3159 ± 0.0154 & 0.4758 ± 0.0103 \\

CBraMod & 25.1M/19.1M & 0.6160 ± 0.0387 & 0.5272 ± 0.0769 & 0.7487 ± 0.0502 & 0.4092 ± 0.0221 & 0.2123 ± 0.0294 & 0.3417 ± 0.0359 \\
\hline
STAMP & 0.72M/0.72M & 0.6438 ± 0.0728 & 0.5889 ± 0.0525 & 0.8114 ± 0.0206 & 0.5564 ± 0.0212 & 0.4086 ± 0.0282 & 0.5512 ± 0.0242 \\
\hline
\end{tabular}
}
\label{tab:mental_arithmetic_bcic}
\end{table*}

\figureref{fig:final_eval_2x2} presents a comparison between STAMP, EEGFMs and non-foundation model EEG baselines. We see that STAMP yields similar or better performance compared to CBraMod and LaBraM across all datasets, while strictly outperforming all other methods. In \tableref{tab:shu_phys,tab:mental_arithmetic_bcic}, the explicit performance metrics for each of the 4 datasets are shown. STAMP uses an average of 0.74M parameters while CBraMod uses an average of 29M and LaBraM uses 5.8M. Despite the model size difference, STAMP frequently outperforms both fully supervised model baselines and achieves overlapping or better performance with the EEGFMs. Further analysis demonstrated that STAMP can often provide the same level of performance with even fewer parameters (see Appendix \ref{apd:stamp_size_comparison}).

We also provide a performance comparison between each method on 4 additional datasets (see \figureref{fig:final_eval_2} and \tableref{tab:tuev_mumtaz,tab:seedv_faced}). STAMP continues to show strong performance, especially on TUEV and Mumtaz2016, where it matches each EEGFM and outperforms non-foundation model EEG baselines. Mumtaz2016 is generally an easier to model dataset, given all models yield near perfect classification, so that result is less significant. However, the result for TUEV points to MOMENT and STAMP's ability to extract and classify clinically-relevant EEG events. For SEED-V and FACED, STAMP yields inferior performance compared to the other methods. Both of those datasets are emotion recognition tasks, which implies that MOMENT may not be able to extract features relevant to the task of emotion recognition, resulting in poor downstream performance with STAMP. Due to this finding, we evaluated STAMP on these two emotion recognition datasets using embeddings from Chronos Large. We found that Chronos provided a small performance boost, yet performance on FACED was still lackluster (see Appendix \ref{apd:emotion_recognition}). 

\subsection{TSFM Comparison}

The comparison of TSFMs demonstrates that STAMP can achieve strong results when used on top of varying TSFMs. \figureref{fig:tsfm_comparison_2x2} shows that for SHU-MI and PhysioNet-MI, each TSFM yields similar performance. Interestingly, MOMENT Small performs just as well as MOMENT Large on three out of the four datasets, while using only $\approx 10\%$ of the parameters, significantly reducing embedding generation time. TSPulse performs worse for the MentalArithmetic and BCIC-IV-2a datasets, however, its performance is exceptional for the other datasets given its small size. Also of note, we found that aggregating output embeddings from Chronos using mean pooling, instead of only using the end of sequence (EOS) embedding, provided stronger downstream performance with STAMP (see Appendix \ref{apd:chronos}).

Each variant of MOMENT pretrains on diverse time series data, a very small portion of which is EEG data. However, to our knowledge, TSPulse and Chronos do not include any EEG data in their pretraining. Thus, these results demonstrate that TSFMs can extract meaningful features from EEG signals, regardless of whether the TSFM was pretrained on any EEG. We are optimistic that performance achievable with STAMP will improve as TSFMs continue to advance.

\begin{figure}[htbp]
\floatconts
  {fig:tsfm_comparison_2x2}
  {\caption{Performance comparison between using the following TSFMs with STAMP: TSPulse (1M, 0.63M), MOMENT Small (40M, 0.67M), MOMENT Base (125M, 0.7M), MOMENT Large (385M, 0.74M), and Chronos Large (710M, 0.74M). The first value in the parentheses indicates the size of the TSFM and the second value denotes the average number of trainable parameters in STAMP across the 4 datasets.}}
  {\includegraphics[width=\linewidth]{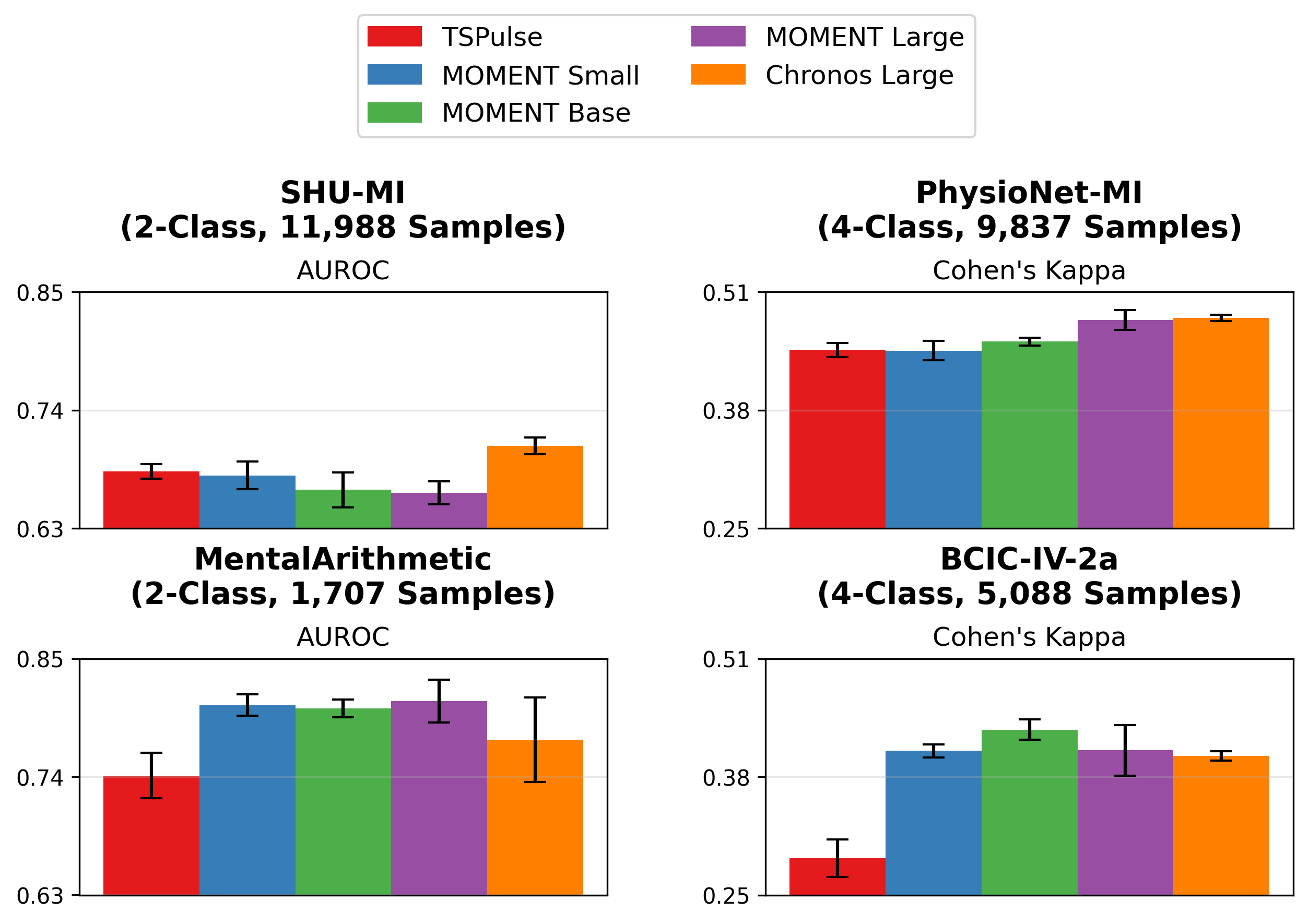}}
\end{figure}

\vspace{-3ex}
\section{Conclusion and Future Work}
\label{sec:discussion}

We present the first spatial-temporal adapter, \textbf{STAMP}, to be used on top of TSFMs for the modeling of EEG data. We demonstrate that the adapter works with multiple TSFMs, performing comparably to EEGFMs with a fraction of the trainable parameters. Throughout the development of the adapter, we learned valuable lessons on employing foundation models for EEG modeling. 

Using the TSFM MOMENT out of the box, with only the addition of mean pooling, resulted in near-random performance (see \figureref{fig:moment_baseline_vs_stamp}). As shown in our ablations, both positional encodings and a token mixing strategy were necessary for generating good predictions with MOMENT embeddings. Although our more advanced token aggregation strategy MHAP performed mostly similarly to mean pooling, better results on the BCIC-IV-2a dataset suggest that our adapter may have more robust performance across datasets when using learnable aggregation schemes.

In summary, without additional modeling to capture relationships between channels and the relative importance of different time points, TSFMs may seem to underperform compared to EEGFMs. However, relatively few adaptations and additional trainable parameters are needed to achieve results competitive with what can be achieved with large-scale EEG pretraining and fine-tuning as in~\cite{wang_cbramod_2025}. While there are added computational resources needed to generate the initial embeddings for downstream use with STAMP, this can be done upfront and in parallel.

Further exploration of adapter performance using other TSFMs in conjunction with TSFM fine-tuning would be of interest (we evaluate only one LoRA configuration), particularly on datasets where STAMP struggled. However, we also found it encouraging that performance benefits were minimal: MOMENT is able to capture representations of EEG data sufficient for various downstream tasks even without adjustment. Other future work of interest could explore STAMP from an interpretability perspective, for example by examining the relative contributions of different tokens to final predictions. Exploring strategies for explicitly incorporating geometric or topographical relationships between electrodes is also a potentially interesting avenue. STAMP in its current form has the advantage of being relatively agnostic to the specifics of EEG data, and so evaluating performance on additional modeling tasks requiring multivariate time-series (e.g. vital signs, imaging, medication dosages) would also be of interest.

\section*{Acknowledgements}
\label{sec:acknowledgments}

This work has been partially supported by the National Institutes of Health (award R01NS124642) and by the National Science Foundation (awards 2406231, 2427948).

\bibliography{references}

@inproceedings{india_self_2019,
	title = {Self Multi-Head Attention for Speaker Recognition},
	url = {https://www.isca-archive.org/interspeech_2019/india19_interspeech.html},
    year = {2019},
	doi = {10.21437/Interspeech.2019-2616},
	eventtitle = {Interspeech 2019},
	pages = {4305--4309},
	booktitle = {Interspeech 2019},
	publisher = {{ISCA}},
	author = {India, Miquel and Safari, Pooyan and Hernando, Javier},
	urldate = {2025-09-08},
	date = {2019-09-15},
	langid = {english},
}

@inproceedings{liu_pay_2021,
	title = {Pay Attention to {MLPs}},
    year = {2021},
	volume = {34},
	url = {https://papers.neurips.cc/paper_files/paper/2021/file/4cc05b35c2f937c5bd9e7d41d3686fff-Paper.pdf},
	pages = {9204--9215},
	booktitle = {Advances in Neural Information Processing Systems},
	publisher = {Curran Associates, Inc.},
	author = {Liu, Hanxiao and Dai, Zihang and So, David and Le, Quoc V},
	urldate = {2025-09-08},
	date = {2021},
}

@inproceedings{zhao_multi-query_2022,
	author={Zhao, Miao and Ma, Yufeng and Ding, Yiwei and Zheng, Yu and Liu, Min and Xu, Minqiang},
  booktitle={ICASSP 2022 - 2022 IEEE International Conference on Acoustics, Speech and Signal Processing (ICASSP)}, 
  title={Multi-Query Multi-Head Attention Pooling and Inter-Topk Penalty for Speaker Verification}, 
  year={2022},
  volume={},
  number={},
  pages={6737-6741},
  doi={10.1109/ICASSP43922.2022.9746178}
}

@inproceedings{wang_cbramod_2025,
    title={{CB}raMod: A Criss-Cross Brain Foundation Model for {EEG} Decoding},
    author={Jiquan Wang and Sha Zhao and Zhiling Luo and Yangxuan Zhou and Haiteng Jiang and Shijian Li and Tao Li and Gang Pan},
    booktitle={The Thirteenth International Conference on Learning Representations},
    year={2025},
    url={https://openreview.net/forum?id=NPNUHgHF2w}
}

@inproceedings{vaswani_attention_2017,
	title = {Attention is All you Need},
    year = {2017},
	volume = {30},
	url = {https://proceedings.neurips.cc/paper_files/paper/2017/hash/3f5ee243547dee91fbd053c1c4a845aa-Abstract.html},
	booktitle = {Advances in Neural Information Processing Systems},
	publisher = {Curran Associates, Inc.},
	author = {Vaswani, Ashish and Shazeer, Noam and Parmar, Niki and Uszkoreit, Jakob and Jones, Llion and Gomez, Aidan N and Kaiser, Ł ukasz and Polosukhin, Illia},
	urldate = {2025-09-08},
	date = {2017}
}

@article{
ansari_chronos_2024,
title={Chronos: Learning the Language of Time Series},
author={Abdul Fatir Ansari and Lorenzo Stella and Ali Caner Turkmen and Xiyuan Zhang and Pedro Mercado and Huibin Shen and Oleksandr Shchur and Syama Sundar Rangapuram and Sebastian Pineda Arango and Shubham Kapoor and Jasper Zschiegner and Danielle C. Maddix and Hao Wang and Michael W. Mahoney and Kari Torkkola and Andrew Gordon Wilson and Michael Bohlke-Schneider and Bernie Wang},
journal={Transactions on Machine Learning Research},
issn={2835-8856},
year={2024},
url={https://openreview.net/forum?id=gerNCVqqtR},
note={Expert Certification}
}

@inproceedings{rasul_lag-llama_2023,
	title={Lag-Llama: Towards Foundation Models for Time Series Forecasting},
    author={Kashif Rasul and Arjun Ashok and Andrew Robert Williams and Arian Khorasani and George Adamopoulos and Rishika Bhagwatkar and Marin Bilo{\v{s}} and Hena Ghonia and Nadhir Hassen and Anderson Schneider and Sahil Garg and Alexandre Drouin and Nicolas Chapados and Yuriy Nevmyvaka and Irina Rish},
    booktitle={R0-FoMo:Robustness of Few-shot and Zero-shot Learning in Large Foundation Models},
    year={2023},
    url={https://openreview.net/forum?id=jYluzCLFDM}
}

@InProceedings{das_decoder-only_2024,
  title = 	 {A decoder-only foundation model for time-series forecasting},
  author =       {Das, Abhimanyu and Kong, Weihao and Sen, Rajat and Zhou, Yichen},
  booktitle = 	 {Proceedings of the 41st International Conference on Machine Learning},
  pages = 	 {10148--10167},
  year = 	 {2024},
  editor = 	 {Salakhutdinov, Ruslan and Kolter, Zico and Heller, Katherine and Weller, Adrian and Oliver, Nuria and Scarlett, Jonathan and Berkenkamp, Felix},
  volume = 	 {235},
  series = 	 {Proceedings of Machine Learning Research},
  month = 	 {21--27 Jul},
  publisher =    {PMLR},
  url = 	 {https://proceedings.mlr.press/v235/das24c.html}
}

@InProceedings{goswami_moment_2024,
  title = 	 {{MOMENT}: A Family of Open Time-series Foundation Models},
  author =       {Goswami, Mononito and Szafer, Konrad and Choudhry, Arjun and Cai, Yifu and Li, Shuo and Dubrawski, Artur},
  booktitle = 	 {Proceedings of the 41st International Conference on Machine Learning},
  pages = 	 {16115--16152},
  year = 	 {2024},
  editor = 	 {Salakhutdinov, Ruslan and Kolter, Zico and Heller, Katherine and Weller, Adrian and Oliver, Nuria and Scarlett, Jonathan and Berkenkamp, Felix},
  volume = 	 {235},
  series = 	 {Proceedings of Machine Learning Research},
  month = 	 {21--27 Jul},
  publisher =    {PMLR},
  url = 	 {https://proceedings.mlr.press/v235/goswami24a.html}
}

@inproceedings{
wang_eegpt_2024,
title={{EEGPT}: Pretrained Transformer for Universal and Reliable Representation of {EEG} Signals},
author={Guangyu Wang and Wenchao Liu and Yuhong He and Cong Xu and Lin Ma and Haifeng Li},
booktitle={The Thirty-eighth Annual Conference on Neural Information Processing Systems},
year={2024},
url={https://openreview.net/forum?id=lvS2b8CjG5}
}

@INPROCEEDINGS{cui_neuro-gpt_2024,
  author={Cui, Wenhui and Jeong, Woojae and Thölke, Philipp and Medani, Takfarinas and Jerbi, Karim and Joshi, Anand A. and Leahy, Richard M.},
  booktitle={2024 IEEE International Symposium on Biomedical Imaging (ISBI)}, 
  title={Neuro-GPT: Towards A Foundation Model For EEG}, 
  year={2024},
  volume={},
  number={},
  pages={1-5},
  doi={10.1109/ISBI56570.2024.10635453}}

@inproceedings{
jiang_large_2024,
title={Large Brain Model for Learning Generic Representations with Tremendous {EEG} Data in {BCI}},
author={Wei-Bang Jiang and Li-Ming Zhao and Bao-Liang Lu},
booktitle={The Twelfth International Conference on Learning Representations},
year={2024},
url={https://openreview.net/forum?id=QzTpTRVtrP}
}

@article{craik_deep_2019,
	title = {Deep learning for electroencephalogram ({EEG}) classification tasks: a review},
    year = {2019},
	volume = {16},
	issn = {1741-2552},
	url = {https://dx.doi.org/10.1088/1741-2552/ab0ab5},
	doi = {10.1088/1741-2552/ab0ab5},
	shorttitle = {Deep learning for electroencephalogram ({EEG}) classification tasks},
	pages = {031001},
	number = {3},
	journaltitle = {Journal of Neural Engineering},
	shortjournal = {J. Neural Eng.},
	author = {Craik, Alexander and He, Yongtian and Contreras-Vidal, Jose L},
	urldate = {2025-08-26},
	date = {2019-04},
	langid = {english},
	note = {Publisher: {IOP} Publishing}
}

@article{liu_seedv_2022,
	author={Liu, Wei and Qiu, Jie-Lin and Zheng, Wei-Long and Lu, Bao-Liang},
  journal={IEEE Transactions on Cognitive and Developmental Systems}, 
  title={Comparing Recognition Performance and Robustness of Multimodal Deep Learning Models for Multimodal Emotion Recognition}, 
  year={2022},
  volume={14},
  number={2},
  pages={715-729},
  doi={10.1109/TCDS.2021.3071170}
}

@article{chen_faced_2023,
	title = {A Large Finer-grained Affective Computing {EEG} Dataset},
    year = {2023},
	volume = {10},
	rights = {2023 The Author(s)},
	issn = {2052-4463},
	url = {https://www.nature.com/articles/s41597-023-02650-w},
	doi = {10.1038/s41597-023-02650-w},
	pages = {740},
	number = {1},
	journaltitle = {Scientific Data},
	shortjournal = {Sci Data},
	author = {Chen, Jingjing and Wang, Xiaobin and Huang, Chen and Hu, Xin and Shen, Xinke and Zhang, Dan},
	urldate = {2025-09-04},
	date = {2023-10-25},
	langid = {english},
	note = {Publisher: Nature Publishing Group},
}

@article{goldberger_physiobank_2000,
	title = {{PhysioBank}, {PhysioToolkit}, and {PhysioNet}: components of a new research resource for complex physiologic signals},
    year = {2000},
	volume = {101},
	issn = {1524-4539},
	doi = {10.1161/01.cir.101.23.e215},
	shorttitle = {{PhysioBank}, {PhysioToolkit}, and {PhysioNet}},
	pages = {E215--220},
	number = {23},
	journaltitle = {Circulation},
	shortjournal = {Circulation},
	author = {Goldberger, A. L. and Amaral, L. A. and Glass, L. and Hausdorff, J. M. and Ivanov, P. C. and Mark, R. G. and Mietus, J. E. and Moody, G. B. and Peng, C. K. and Stanley, H. E.},
	date = {2000-06-13},
	pmid = {10851218},
}

@article{ma_shu_2022,
	title = {A large {EEG} dataset for studying cross-session variability in motor imagery brain-computer interface},
    year = {2022},
	volume = {9},
	rights = {2022 The Author(s)},
	issn = {2052-4463},
	url = {https://www.nature.com/articles/s41597-022-01647-1},
	doi = {10.1038/s41597-022-01647-1},
	pages = {531},
	number = {1},
	journaltitle = {Scientific Data},
	shortjournal = {Sci Data},
	author = {Ma, Jun and Yang, Banghua and Qiu, Wenzheng and Li, Yunzhe and Gao, Shouwei and Xia, Xinxing},
	urldate = {2025-09-04},
	date = {2022-09-01},
	langid = {english},
	note = {Publisher: Nature Publishing Group},
}

@article{schalk_bci2000_2004,
	author={Schalk, Gerwin and McFarland, Dennis J and Hinterberger, Thilo and Birbaumer, Niels and Wolpaw, Jonathan R},
  journal={IEEE Transactions on Biomedical Engineering}, 
  title={BCI2000: a general-purpose brain-computer interface (BCI) system}, 
  year={2004},
  volume={51},
  number={6},
  pages={1034-1043},
  doi={10.1109/TBME.2004.827072}}

@article{mumtaz_mdd_2016,
	title = {{MDD} Patients and Healthy Controls {EEG} Data (New)},
    year = {2016},
	url = {https://figshare.com/articles/dataset/EEG_Data_New/4244171},
	doi = {10.6084/m9.figshare.4244171.v2},
	author = {Mumtaz, Wajid},
	date = {2016-11},
}

@article{zyma_stress_2019,
	title = {Electroencephalograms during Mental Arithmetic Task Performance},
    year = {2019},
	volume = {4},
	rights = {http://creativecommons.org/licenses/by/3.0/},
	issn = {2306-5729},
	url = {https://www.mdpi.com/2306-5729/4/1/14},
	doi = {10.3390/data4010014},
	pages = {14},
	number = {1},
	journaltitle = {Data},
	author = {Zyma, Igor and Tukaev, Sergii and Seleznov, Ivan and Kiyono, Ken and Popov, Anton and Chernykh, Mariia and Shpenkov, Oleksii},
	urldate = {2025-09-04},
	date = {2019-03},
	langid = {english},
	note = {Publisher: Multidisciplinary Digital Publishing Institute},
}

@article{obeid_tuh_2016,
AUTHOR={Obeid, Iyad  and Picone, Joseph },
TITLE={The Temple University Hospital EEG Data Corpus},
JOURNAL={Frontiers in Neuroscience},
VOLUME={Volume 10 - 2016},
YEAR={2016},
URL={https://www.frontiersin.org/journals/neuroscience/articles/10.3389/fnins.2016.00196},
DOI={10.3389/fnins.2016.00196},
ISSN={1662-453X},

}

@article{brunner_bci2a_nodate,
  author  = {Brunner, Clemens and Leeb, Robert and Müller-Putz, Gernot and Schlögl, Alois and Pfurtscheller, Gert},
  title   = {BCI Competition 2008 -- Graz Data Set A},
  journal = {Institute for Knowledge Discovery (Laboratory of Brain-Computer Interfaces), Graz University of Technology},
  volume  = {16},
  pages   = {1--6},
  year    = {2008},
}

@inproceedings{li_seedv_2019,
	location = {San Francisco, {CA}, {USA}},
    year = {2019},
	title = {Classification of Five Emotions from {EEG} and Eye Movement Signals: Discrimination Ability and Stability over Time},
	rights = {https://ieeexplore.ieee.org/Xplorehelp/downloads/license-information/{IEEE}.html},
	isbn = {978-1-5386-7921-0},
	url = {https://ieeexplore.ieee.org/document/8716943/},
	doi = {10.1109/NER.2019.8716943},
	shorttitle = {Classification of Five Emotions from {EEG} and Eye Movement Signals},
	eventtitle = {2019 9th International {IEEE}/{EMBS} Conference on Neural Engineering ({NER})},
	pages = {607--610},
	booktitle = {2019 9th International {IEEE}/{EMBS} Conference on Neural Engineering ({NER})},
	publisher = {{IEEE}},
	author = {Li, Tian-Hao and Liu, Wei and Zheng, Wei-Long and Lu, Bao-Liang},
	urldate = {2025-09-06},
	date = {2019-03},
	langid = {english},
}

@article{harati_tuev_2015,
	title = {Improved {EEG} Event Classification Using Differential Energy},
    year = {2015},
	volume = {2015},
	issn = {2372-7241},
	url = {https://www.ncbi.nlm.nih.gov/pmc/articles/PMC4874511/},
	doi = {10.1109/SPMB.2015.7405421},
	pages = {10.1109/SPMB.2015.7405421},
	journaltitle = {... {IEEE} Signal Processing in Medicine and Biology Symposium. {IEEE} Signal Processing in Medicine and Biology Symposium},
	shortjournal = {{IEEE} Signal Process Med Biol Symp},
	author = {Harati, A. and Golmohammadi, M. and Lopez, S. and Obeid, I. and Picone, J.},
	urldate = {2025-09-06},
	date = {2015-12},
	pmid = {27213180},
	pmcid = {PMC4874511},
}

@article{hu2022lora,
  title={Lora: Low-rank adaptation of large language models.},
  author={Hu, Edward J and Shen, Yelong and Wallis, Phillip and Allen-Zhu, Zeyuan and Li, Yuanzhi and Wang, Shean and Wang, Lu and Chen, Weizhu and others},
  journal={ICLR},
  volume={1},
  number={2},
  pages={3},
  year={2022}
}

@ARTICLE{eeg_conformer,
    author={Song, Yonghao and Zheng, Qingqing and Liu, Bingchuan and Gao, Xiaorong},
  journal={IEEE Transactions on Neural Systems and Rehabilitation Engineering}, 
  title={EEG Conformer: Convolutional Transformer for EEG Decoding and Visualization}, 
  year={2023},
  volume={31},
  number={},
  pages={710-719},
  doi={10.1109/TNSRE.2022.3230250}}

@misc{st_transformer,
	title={Transformer-based Spatial-Temporal Feature Learning for EEG Decoding}, 
      author={Yonghao Song and Xueyu Jia and Lie Yang and Longhan Xie},
      year={2021},
      eprint={2106.11170},
      archivePrefix={arXiv},
      primaryClass={eess.SP},
      url={https://arxiv.org/abs/2106.11170}, 
}

@misc{ekambaram2025tspulsedualspacetiny,
      title={TSPulse: Dual Space Tiny Pre-Trained Models for Rapid Time-Series Analysis}, 
      author={Vijay Ekambaram and Subodh Kumar and Arindam Jati and Sumanta Mukherjee and Tomoya Sakai and Pankaj Dayama and Wesley M. Gifford and Jayant Kalagnanam},
      year={2025},
      eprint={2505.13033},
      archivePrefix={arXiv},
      primaryClass={cs.LG},
      url={https://arxiv.org/abs/2505.13033}, 
}

\clearpage
\appendix

\section{Datasets}\label{apd:datasets}

There are a total of 8 datasets that we use for evaluation. Each dataset was preprocessed using the publicly available code from \cite{wang_cbramod_2025}.  The preprocessing resamples the samples to 200Hz and each temporal channel is a duration of 1 second, resulting in patches of length 200. We refer to EEG channels as spatial channels and the 1 second durations as temporal channels. The train, validation, and test splits are the same as \cite{wang_cbramod_2025}. In the following, we provide a brief overview of each dataset. 

\begin{table*}[htbp]
\centering
\caption{Dataset characteristics.}
\resizebox{\textwidth}{!}{%
\begin{tabular}{l c c c c c c c c}
\hline
\textbf{Dataset} & \textbf{Classes} & 
\begin{tabular}{c}\textbf{Spatial}\\\textbf{Channels}\end{tabular} & 
\begin{tabular}{c}\textbf{Temporal}\\\textbf{Channels}\end{tabular} & 
\textbf{Samples} & \textbf{Training} & \textbf{Validation} & \textbf{Test} \\
\hline
SHU-MI & 2 & 32 & 4 & 11,988 & 7,210 & 2,431 & 2,347 \\
MentalArithmetic & 2 & 20 & 5 & 1,707 & 1,343 & 172 & 192 \\
BCIC-IV-2a & 4 & 22 & 4 & 5,088 & 2,784 & 1,152 & 1,152 \\
PhysioNet-MI & 4 & 64 & 4 & 9,837 & 6,300 & 1,734 & 1,803 \\
Mumtaz2016 & 2 & 19 & 5 & 7,143 & 4,891 & 1,041 & 1,211 \\
SEED-V & 5 & 62 & 1 & 117,744 & 34,432 & 42,960 & 40,352 \\
TUEV & 6 & 16 & 5 & 113,353 & 68,445 & 15,487 & 29,421 \\
FACED & 9 & 32 & 10 & 10,332 & 6,720 & 1,680 & 1,932 \\
\hline
\end{tabular}
}
\label{tab:datasets}
\end{table*}

\textbf{SHU-MI}: An EEG dataset containing recordings from a motor imagery task. The task consisted of sitting in a chair in front of an LCD monitor and an image denoting a right-handed or left-handed movement (2 classes) was presented on screen. When the movement was shown, the subject repeatedly imagined the movement. The original dataset description and methodology is provided by \cite{ma_shu_2022}.

\textbf{PhysioNet-MI}: An EEG motor imagery dataset. The task involved a target appearing on a screen and the subject performing one of 4 actions, which serves as our classes: opening and closing a fist, imagining opening and closing a fist, opening and closing both fists or both feet, imagining opening and closing both fists or both feet. For the original dataset description, see \cite{schalk_bci2000_2004}. 

\textbf{MentalArithmetic}: An EEG cognitive dataset involving the study of mental activity while performing a cognitively intensive task, mental arithmetic. Subjects were asked to subtract a 2-digit number from a 4-digit number. The EEG recordings before the task are labeled as "without mental stress" and the recordings during the task are labeled as "with mental stress", resulting in 2 classes. The original dataset description can be found via \cite{zyma_stress_2019}.

\textbf{BCIC-IV-2a} An EEG motor imagery dataset containing 4 movements (4 classes). Each subject sat in front of a screen and a visual cue displayed on the screen indicating which movement to perform. The movements involved one of 4 body parts: left hand, right hand, both feet, and tongue. A detailed description of the dataset is written by \cite{brunner_bci2a_nodate}.

\textbf{SEED-V}: An EEG emotion recognition dataset focusing on the emotional response of subjects after viewing movie clips intended to provoke a specific emotion. There are 5 classes of emotions in the dataset: happy, sad, neutral, fear, and disgust. \cite{li_seedv_2019} describes the dataset collection and preprocessing. 

\textbf{Mumtaz2016}: An EEG mental health dataset concentrating on the identification of major depressive disorder based on EEG. The EEG recordings were performed on normal control patients and patients with major depressive disorder, resulting in 2 classes. Only limited details about the dataset are available~\citep{mumtaz_mdd_2016}. 
 
\textbf{TUEV}: An EEG event type dataset containing EEG signals belonging to one of 6 classes: spike and sharp wave (SPSW), generalized periodic epileptiform discharges (GPED), periodic lateralized epileptiform discharges (PLED), eye movement (EYEM), artifact (ARTF), and background (BCKG). To note, CBraMod reports using 112,491 samples. Despite using their preprocessing code, our preprocessed dataset had a total of 113,353 samples. For a detailed description of the dataset, refer to \cite{harati_tuev_2015, obeid_tuh_2016}. 

\textbf{FACED}: A 9-class EEG emotion recognition dataset. The subjects watched 28 video clips intended to elicit a specific emotion. The emotions included amusement, inspiration, joy, tenderness, anger, fear, disgust, sadness, and neutral emotion. \cite{chen_faced_2023} gives a detailed summary of the collection and preprocessing of the dataset. 

\section{Hyperparameters}\label{apd:hyperparameters}

All STAMP experiments (excluding those with LoRA) used the hyperparameters listed in \tableref{tab:general_stamp_hyperparams}.

\begin{table}[htbp]
\centering
\caption{Hyperparameters used in our experiments.}
\begin{tabular}{l c}
\hline
\textbf{Hyperparameter} & \textbf{Value} \\
\hline
Learning rate scheduler & OneCycle \\
Initial learning rate   & 5e-5 \\
Max learning rate       & 0.0003 \\
Batch size              & 64 \\
Optimizer               & AdamW \\
Epsilon                 & 1e-8 \\
Weight decay            & 0.05 \\
Dropout rate            & 0.3 \\
Epochs                  & 50 \\
D                       & 128 \\
Normalization           & Instance \\
\hline
\end{tabular}
\label{tab:general_stamp_hyperparams}
\end{table}

All of our experiments used $L=8$ and $\text{Feedforward Dimension}=256$, where $L$ denotes the number of blocks in the case of GMLP and both the number of attention heads and layers in transformers. Note the feedforward dimension corresponds to $h$ in our CC-GMLP formulation. In all experiments, MHAP used 4 heads and 8 query vectors per head. The best epoch during training was chosen based on the validation monitor metric and the model weights at that checkpoint were used for evaluation on the test split. 

When LoRA was part of an experiment, alpha was set to 32, rank to 16, and its dropout rate to 0.05. For the LoRA experiments, we were forced to use a smaller batch size (16) due to GPU memory limitations. Additionally, we used 15 epochs, an initial learning rate of $1e-4$, and a max learning rate of $1e-3$. All other hyperparameters were kept the same.

When reproducing the CBraMod results, we used the hyperparameters mentioned by \cite{wang_cbramod_2025} and if any were not mentioned, we set them to the default value used in their code repository\footnote{\url{https://github.com/wjq-learning/CBraMod}}. 

The main seed that we used was 42 and from that we randomly generated seeds 654, 114, 25, 759, and 281. The first three seeds were used for ablation experiments and hyperparameter tuning. All five seeds were used for our full evaluation. As a result of our fixed seeds, each experiment is fully reproducible. 

\section{Hyperparameter Tuning}\label{apd:hp_tuning}

During the development of the adapter, many hyperparameters were searched over. Our final hyperparameters were selected based on performance on the validation splits of the four main datasets. We found that there was not a consistent performance trend between 0.1, 0.3, and 0.5 dropout rate. After determining CC-GMLP yielded stronger performance than the other token mixing alternatives, we searched over combinations of different $D\in\{128, 256, 512, 1024\}$ and CC-GMLP hyperparameters: $L\in\{2, 4, 8\}$ and $\text{Feedforward Dimension}\in\{256, 512\}$. This search demonstrated that $D=128,\ L=8$ and Feedforward Dimension$=256$ provided the best performance to parameter count ratio; increasing the adapter size did not provide a consistent change in performance. 

\section{Size Comparison}\label{apd:stamp_size_comparison}

One hyperparameter that has a high impact on the size of STAMP is $D$, which is the dimension to which the initial TSFM embeddings are projected. During the development of STAMP, we found that using $D = 128$ yielded consistently strong results, while keeping the number of parameters low. However, we additionally observe that STAMP can often perform strongly at lower $D$ values, even as low as $D = 8$. \figureref{fig:d_comparison} shows how performance varies as $D$ changes and the effect of $D$ on the parameter count. For each dataset, except BCIC-IV-2a, STAMP with $D=8$ achieves similar performance as $D=128$ while using on average $\approx 91\%$ less parameters. This finding highlights that, depending on the task, STAMP can be made further lightweight while maintaining impressive performance. To note, each of these experiments used the previously mentioned 5 seeds and all other hyperparameters of STAMP were fixed.

\begin{figure}[htbp]
\floatconts
  {fig:d_comparison}
  {\caption{Performance comparison of STAMP with varying $D$ values.}}
  {\includegraphics[width=\linewidth]{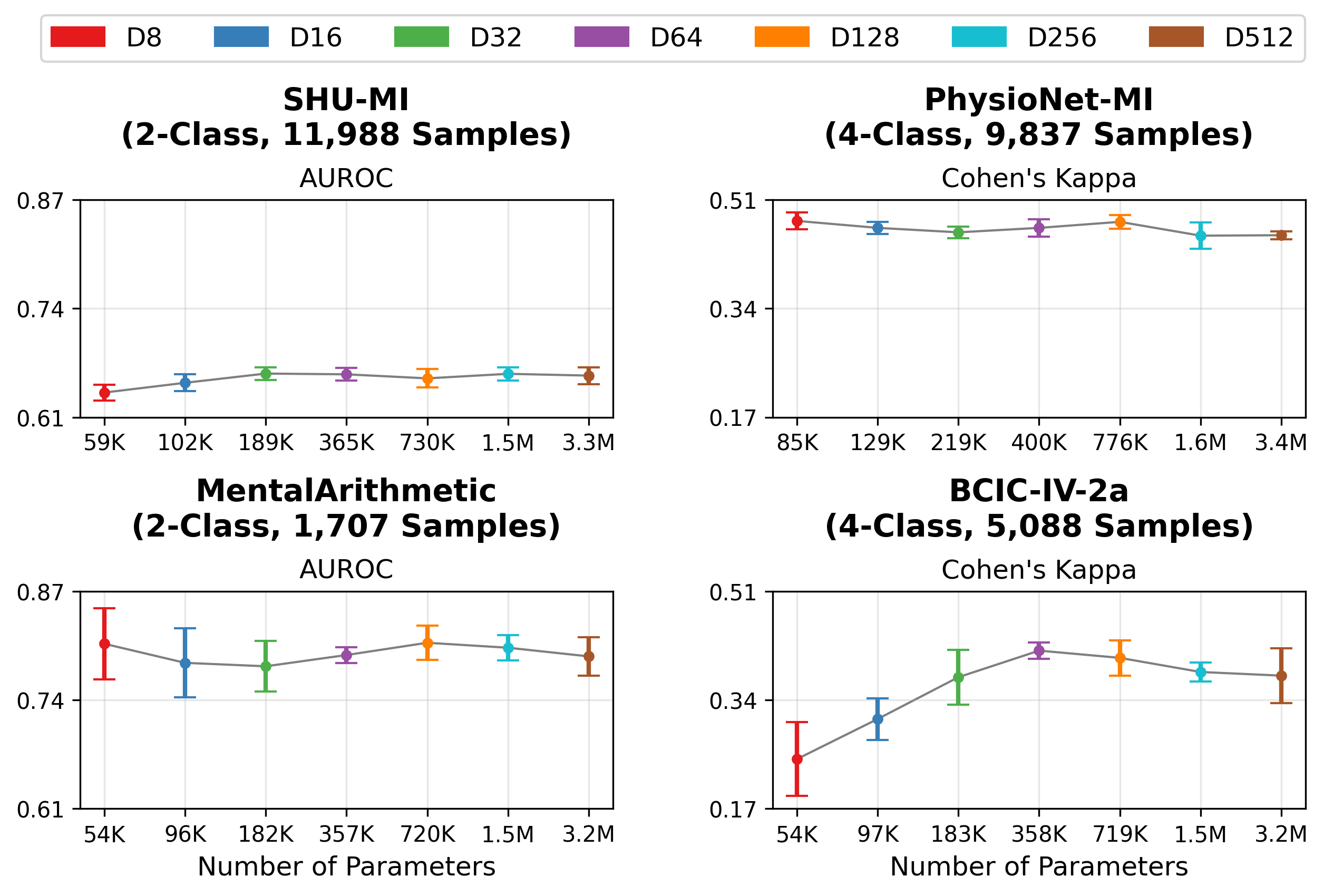}}
\end{figure}

\section{Chronos Embedding Aggregation}\label{apd:chronos}

\begin{figure}[htbp]
\floatconts
  {fig:chronos_aggregation}
  {\caption{Performance comparison of STAMP when using embeddings from Chronos Large with only the EOS embedding versus an embedding from mean pooling.}}
  {\includegraphics[width=\linewidth]{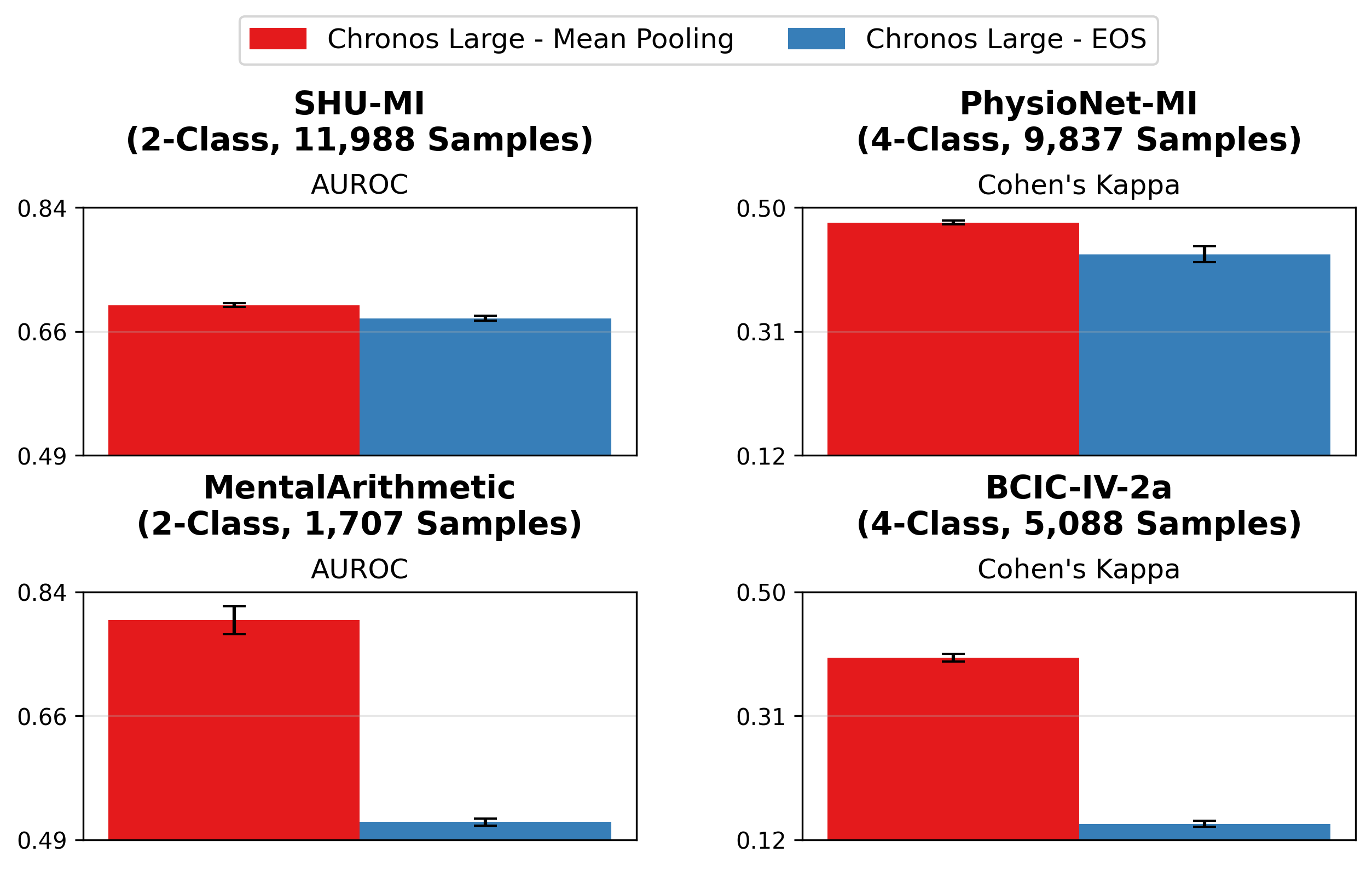}}
\end{figure}

Chronos was primarily built for forecasting and has not been widely adapted for classification. However, embeddings can be extracted from Chronos and used for downstream tasks. When Chronos embeds a time series $x$ with length 200, the output is $e \in \mathbb{R}^{201 \times 1024}$ where 1024 is the embedding dimension for Chronos Large. The first 200 embeddings correspond to the length of the time series and the last embedding corresponds to an EOS token. For use with STAMP, we reduced these embeddings to a single representative embedding. We tested two aggregation methods: 1) mean pooling across all 200 embeddings and 2) using only the EOS embedding. \figureref{fig:chronos_aggregation} shows that using an embedding from mean pooling greatly outperforms only using the EOS embedding. Notably, only the 3 previously mentioned seeds were used for this comparison.

\section{Emotion Recognition Results}\label{apd:emotion_recognition}

\begin{figure}[htbp]
\floatconts
  {fig:emotion_recognition}
  {\caption{Performance comparison of baselines, MOMENT embeddings with STAMP, and Chronos Large embeddings with STAMP on the two emotion recognition datasets: SEED-V and FACED.}}
  {\includegraphics[width=\linewidth]{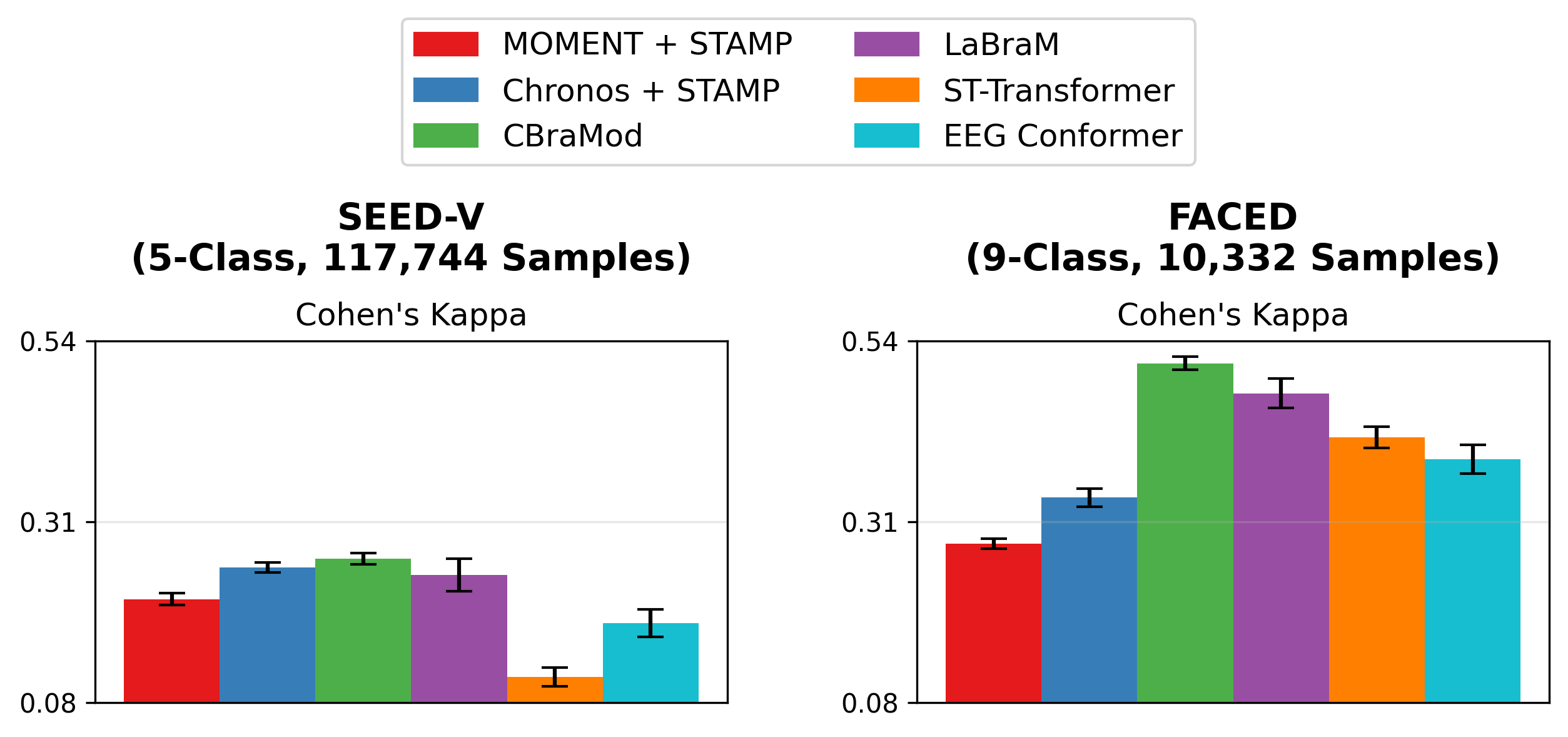}}
\end{figure}

To further analyze the reason that STAMP performs poorly for the two emotion recognition datasets, SEED-V and FACED, we ran an additional STAMP experiment using embeddings from Chronos Large. The previously mentioned 5 seeds were used for this experiment. In \figureref{fig:emotion_recognition}, we see that Chronos + STAMP performs slightly better than MOMENT + STAMP. With Chronos embeddings, STAMP is able to match the performance of the EEGFM baselines on SEED-V, however, the performance on FACED is still poor. Given the comparatively worse performance of both TSFM approaches on FACED, we suspect that current TSFMs may not be able to extract features necessary for distinguishing between the relatively high number of fine-grained classes in FACED.

\section{Temporal Channel Selection Comparison}\label{apd:temporal_channel}

\begin{figure}[htbp]
\floatconts
  {fig:temporal_channel_comparison}
  {\caption{Performance comparison of STAMP using varying numbers of temporal channels.}}
  {\includegraphics[width=\linewidth]{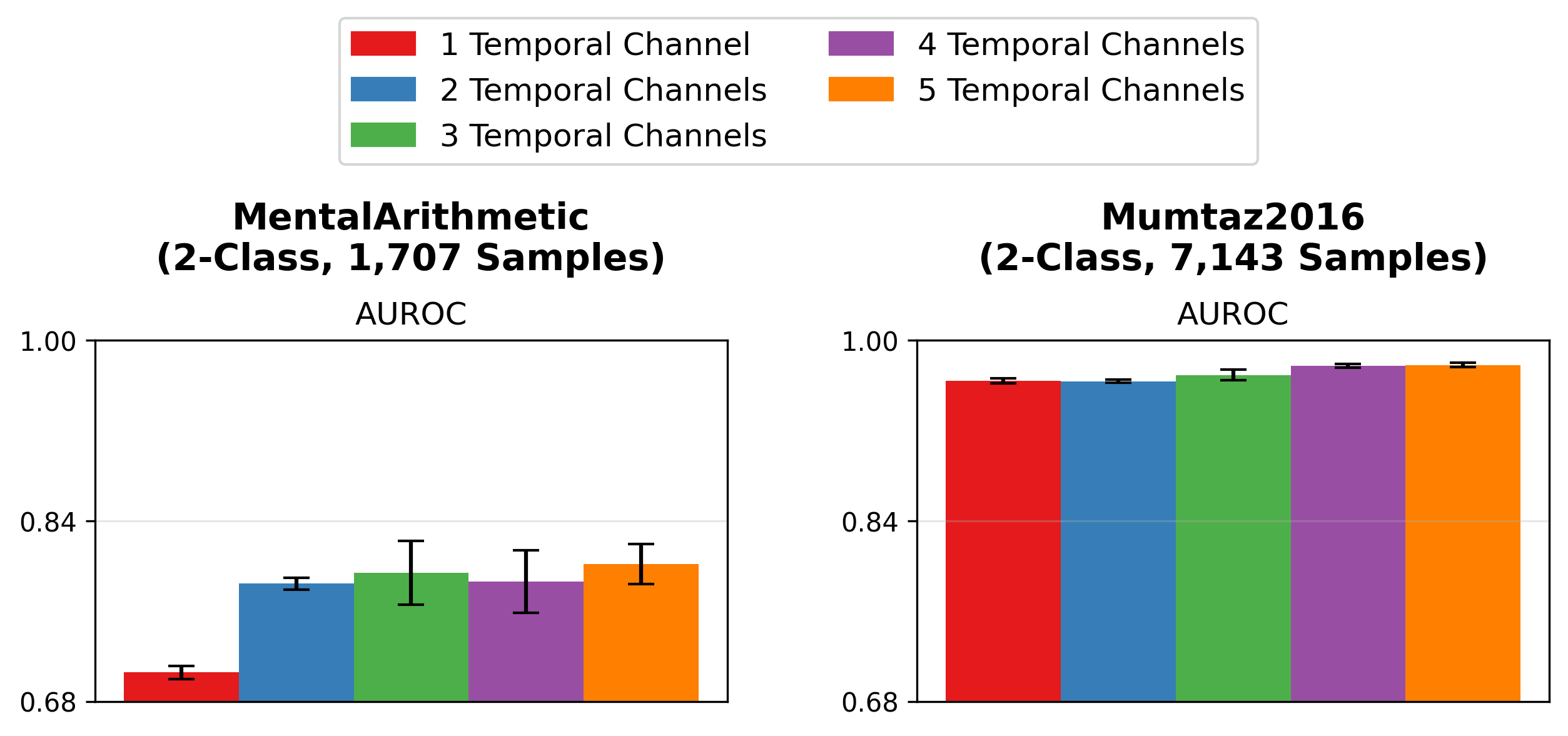}}
\end{figure}

To investigate how the availability of temporal channels affects performance, we performed STAMP experiments using the first $t$ temporal channels, where $t \in \{1,2,3,4,5\}$. In \figureref{fig:temporal_channel_comparison}, we see that performance for MentalArithmetic is similar for all channel selections except the first temporal channel. For Mumtaz2016, temporal channel availability did not greatly affect performance. 

\section{STAMP vs. EEG Conformer}\label{apd:stamp_vs_eeg_conformer}

\begin{figure}[htbp]
\floatconts
  {fig:stamp_vs_eeg_conformer}
  {\caption{Performance comparison of STAMP using $D=96$ versus EEG Conformer. Both methods have $\approx$ 0.55M trainable parameters.}}
  {\includegraphics[width=\linewidth]{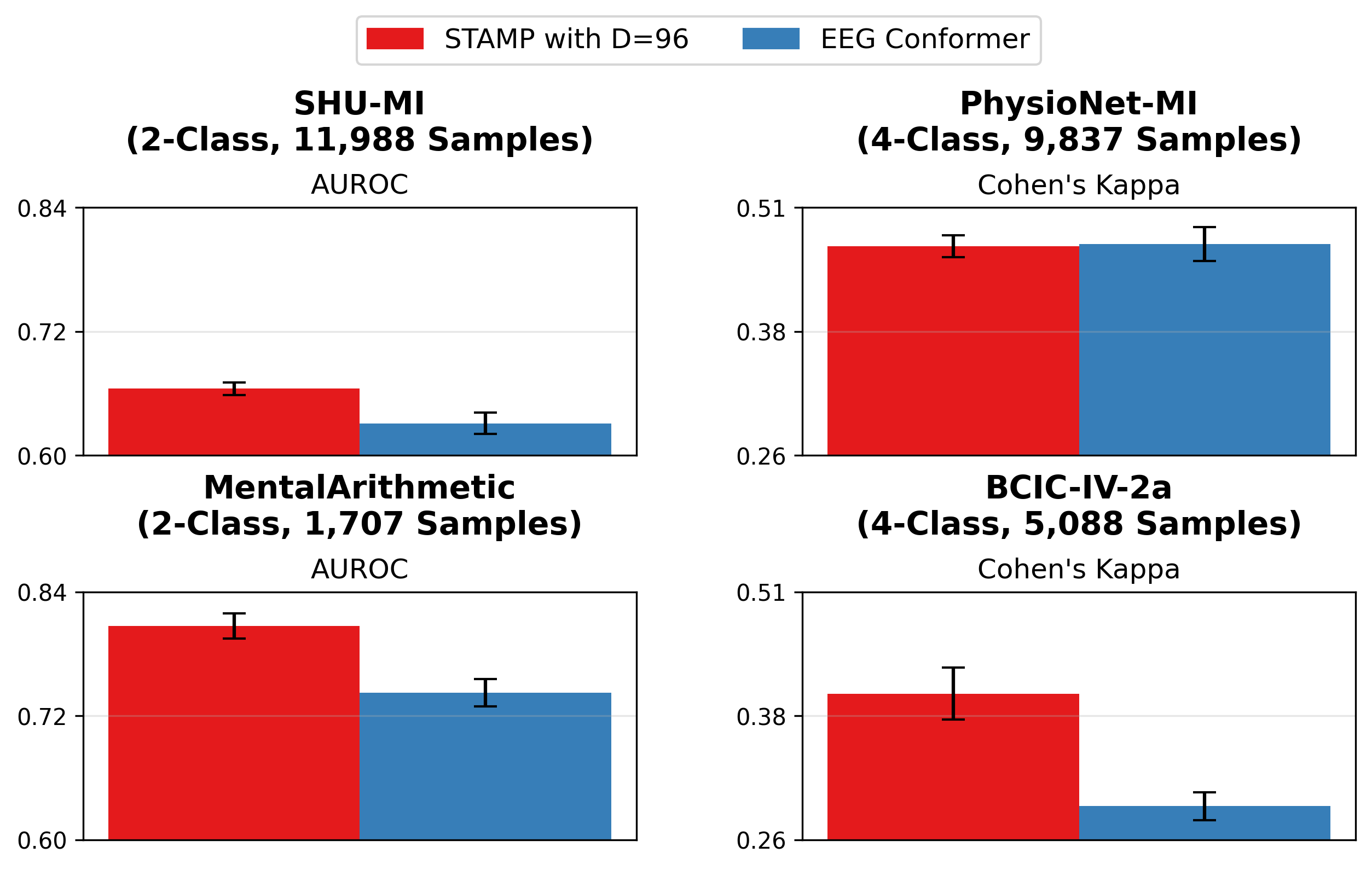}}
\end{figure}

STAMP provides superior performance compared to EEG Conformer for nearly all datasets and metrics evaluated. One may argue that this is due to increased capacity in STAMP and that EEG Conformer is a more efficient method for EEG modeling. In order to demonstrate that this is not the case, we evaluated STAMP using $D=96$ which results in $\approx$ 0.55M parameters matching the parameter count of EEG Conformer. In \figureref{fig:stamp_vs_eeg_conformer}, we see that STAMP outperforms EEG Conformer on 3 of the 4 datasets. As demonstrated by Appendix \ref{apd:stamp_size_comparison}, STAMP can use even fewer parameters and still yield similar performance. 

\section{Additional Results}\label{apd:additional_results}

\begin{figure}[htbp]
\floatconts
  {fig:moment_baseline_vs_stamp}
  {\caption{Performance comparison between MOMENT with mean pooling (0.04M) versus using MOMENT with STAMP (0.74M). The value in parentheses indicates the average number of trainable parameters across the 4 datasets.}}
  {\includegraphics[width=\linewidth]{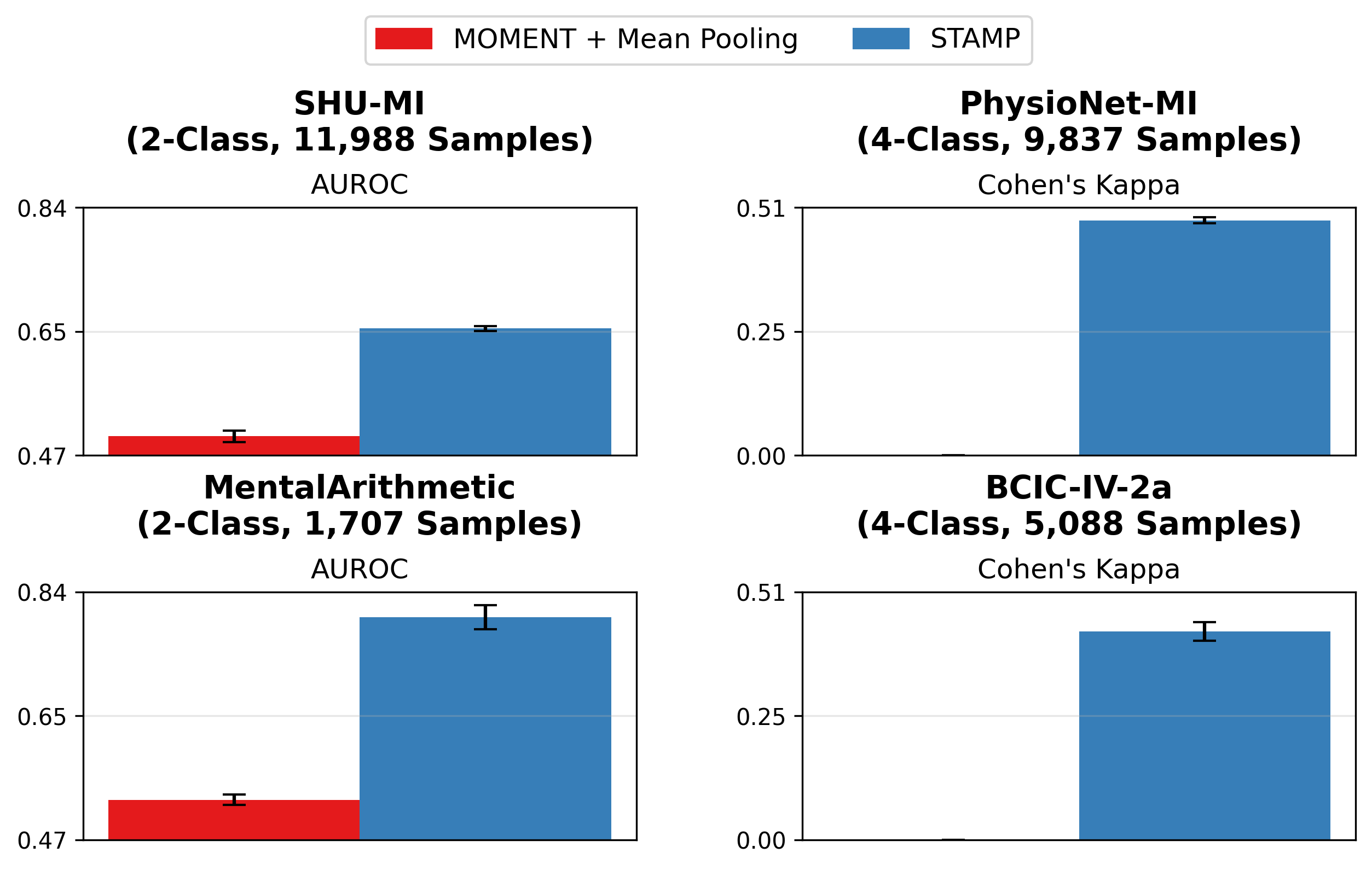}}
\end{figure}

The most naive baseline involving MOMENT is to use only mean pooling on the embeddings. \figureref{fig:moment_baseline_vs_stamp} demonstrates that this naive approach does not learn to model the EEG data and instead results in near-random performance. We see that STAMP provides a significant performance boost compared to this baseline. 

\begin{figure}[htbp]
\floatconts
  {fig:moment_mean_pooling_vs_mhap_no_token_mixing}
  {\caption{Performance comparison between three variants of STAMP: PE-NST + Mean Pooling, PE-NST + MHAP, PE-NST + CC-GMLP + MHAP.}}
  {\includegraphics[width=\linewidth]{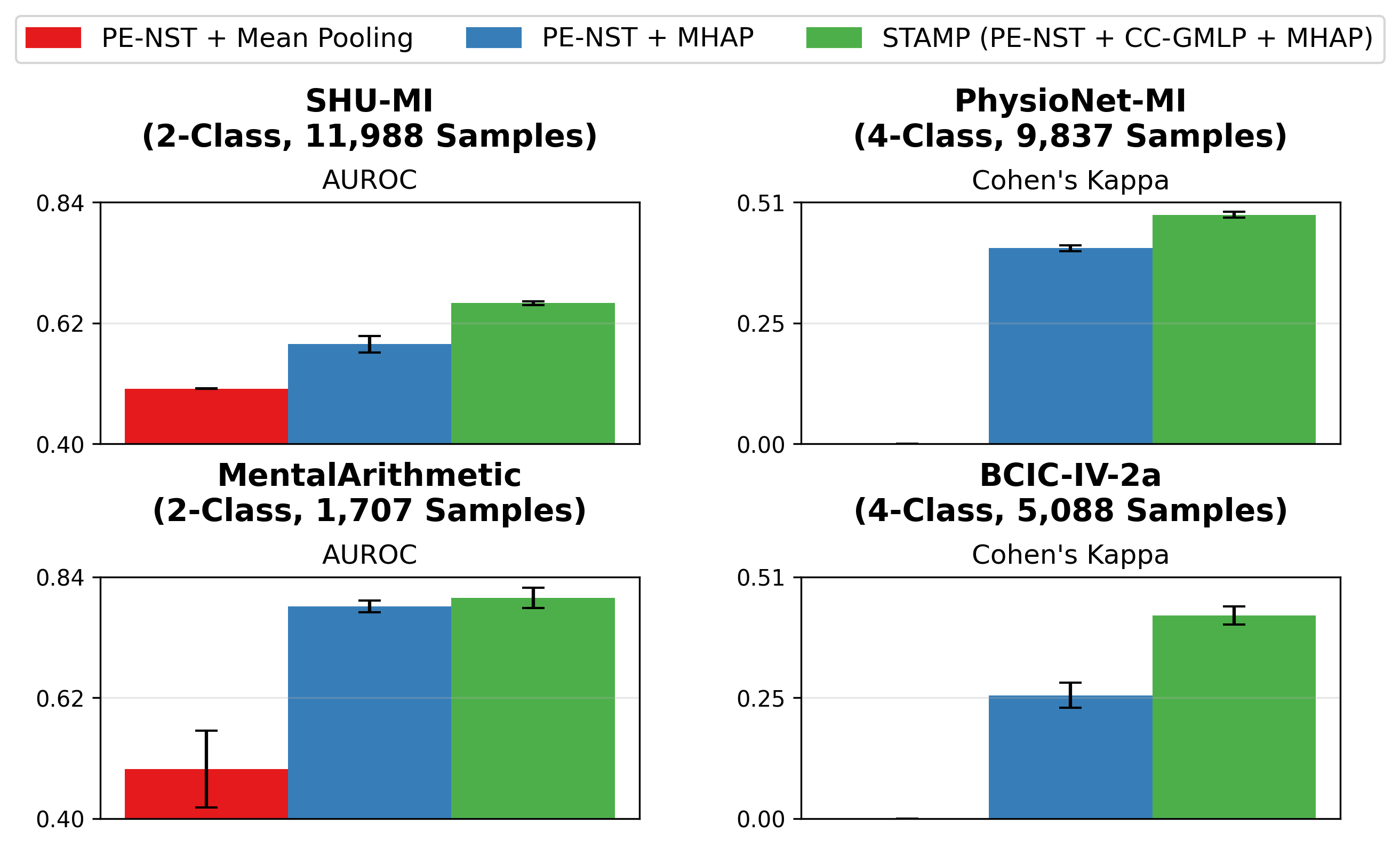}}
\end{figure}

As demonstrated by \figureref{fig:moment_mean_pooling_vs_mhap_no_token_mixing}, PE-NST + mean pooling results in near-random performance without token mixing. However, PE-NST + MHAP is still able to perform reasonably without token mixing. When CC-GMLP is added, further improvement in performance is provided. This comparison demonstrates that the adapter requires some form of relationship modeling in the form of either token mixing or MHAP.

\begin{figure}[htbp]
\floatconts
  {fig:finetune}
  {\caption{Performance comparison between finetuning MOMENT using LoRA and STAMP (2.3M) versus only finetuning STAMP (0.74M). The value in parentheses indicates the average number of trainable parameters across the 4 datasets.}}
  {\includegraphics[width=\linewidth]{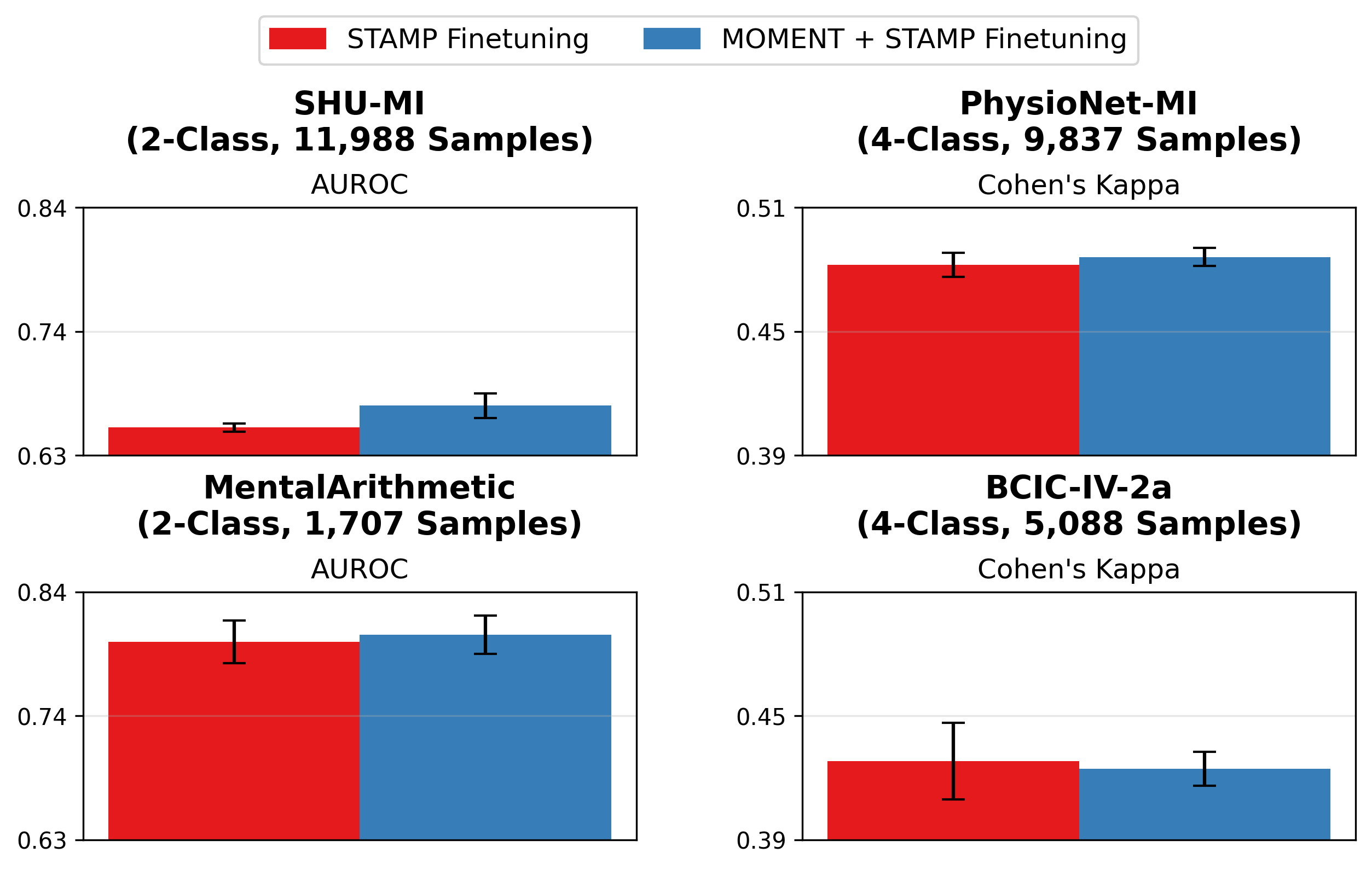}}
\end{figure}

\figureref{fig:finetune} highlights that LoRA does not provide a significant performance increase and the boost that it does provide does not match the additional computational complexity required for fine-tuning. However, we only tested a single configuration of LoRA, so other configurations may provide better results. 

\begin{figure*}[htbp]
\floatconts
  {fig:final_eval_1}
  {\caption{Performance comparison (across all metrics) between the full evaluation of 5 methods: STAMP, CBraMod, LaBraM, ST-Transformer, and EEG Conformer on SHU-MI, PhysioNet-MI, MentalArithmetic, and BCIC-IV-2a.}}
  {\includegraphics[width=\linewidth]{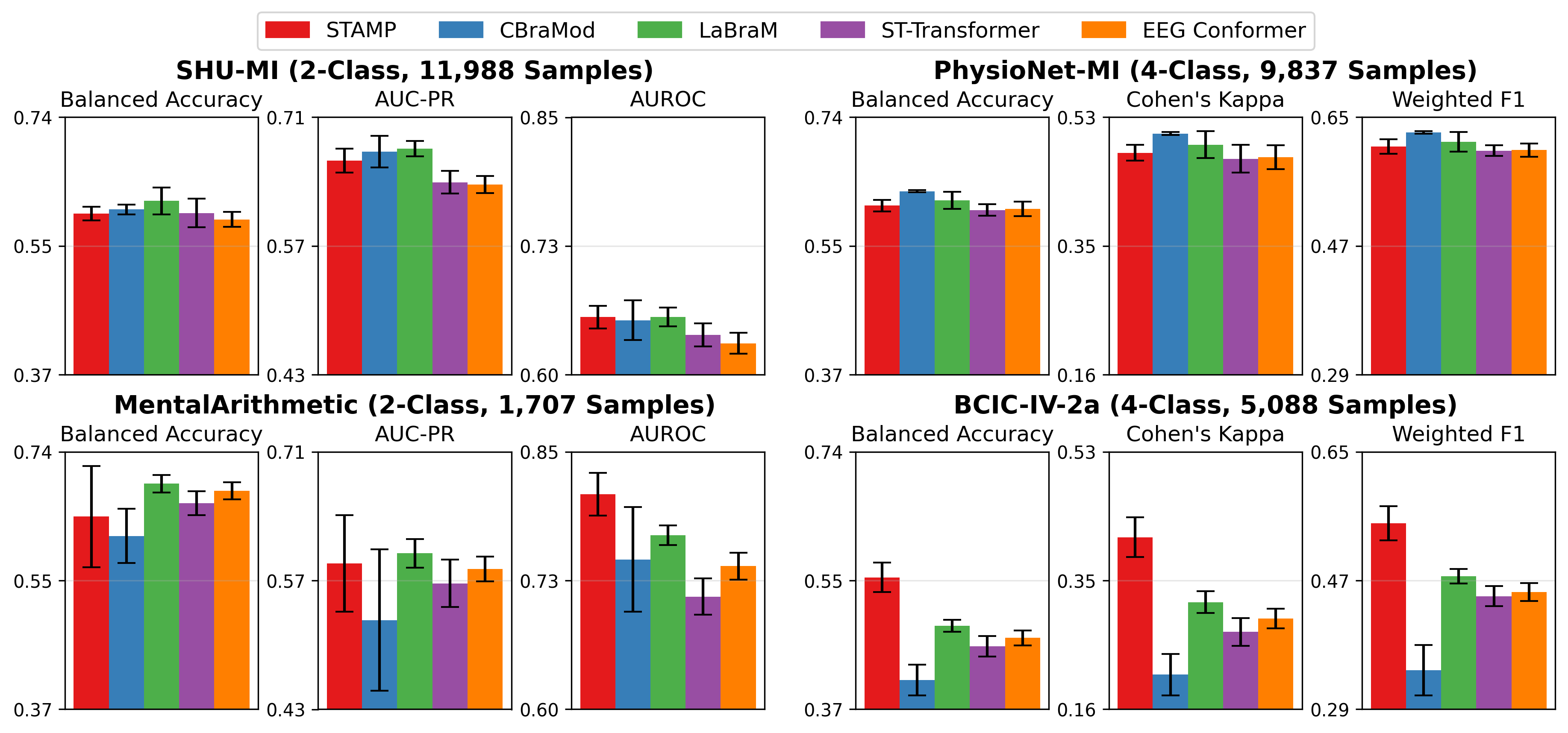}}
\end{figure*}

\begin{figure*}[htbp]
\floatconts
  {fig:final_eval_2}
  {\caption{Performance comparison (across all metrics) between the full evaluation of 5 methods: STAMP, CBraMod, LaBraM, ST-Transformer, and EEG Conformer on TUEV, Mumtaz2016, SEED-V, and FACED.}}
  {\includegraphics[width=\linewidth]{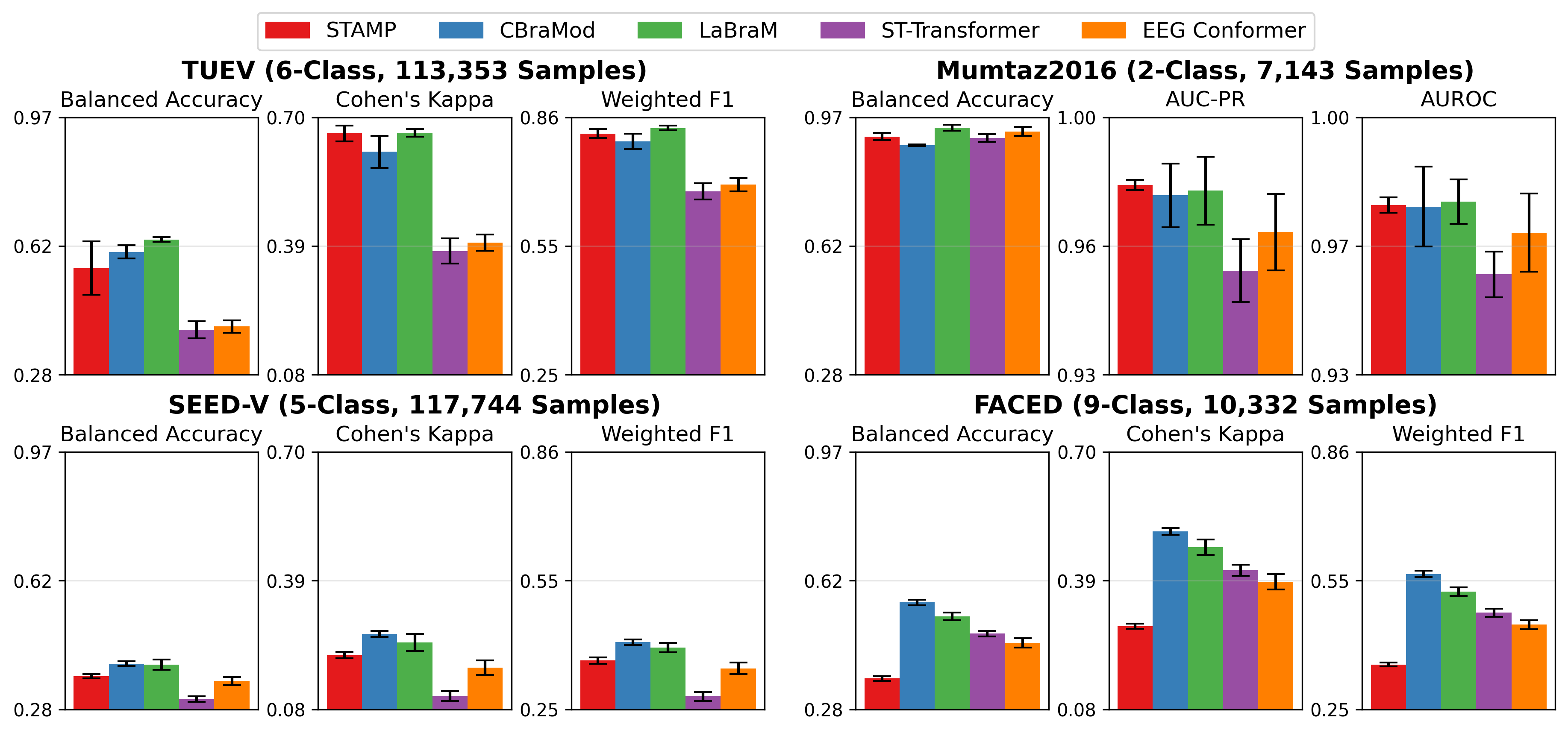}}
\end{figure*}

\begin{table*}[htbp]
\centering
\caption{Performance comparison of different methods on TUEV and Mumtaz2016 datasets.}
\resizebox{\textwidth}{!}{%
\begin{tabular}{|l|r|ccc|ccc|}
\hline
 & & \multicolumn{3}{c|}{\textbf{TUEV (6-Class, 113,353 Samples)}} & \multicolumn{3}{c|}{\textbf{Mumtaz2016 (2-Class, 7,143 Samples)}} \\
\cline{3-8}
\textbf{Methods} & \textbf{\#Params} & \textbf{Balanced Acc.} & \textbf{Cohen's Kappa} & \textbf{Weighted F1} & \textbf{Balanced Acc.} & \textbf{AUC-PR} & \textbf{AUROC} \\
\hline
EEG Conformer & 0.55M & 0.4074 ± 0.0164 & 0.3967 ± 0.0195 & 0.6983 ± 0.0152 & 0.9308 ± 0.0117 & 0.9684 ± 0.0105 & 0.9702 ± 0.0101 \\

ST-Transformer & 3.5M & 0.3984 ± 0.0228 & 0.3765 ± 0.0306 & 0.6823 ± 0.0190 & 0.9135 ± 0.0103 & 0.9578 ± 0.0086 & 0.9594 ± 0.0059 \\
\hline
LaBraM & 5.8M & 0.6409 ± 0.0065 & 0.6637 ± 0.0093 & 0.8312 ± 0.0052 & 0.9409 ± 0.0079 & 0.9798 ± 0.0093 & 0.9782 ± 0.0057 \\

CBraMod & 21.1M/24.1M & 0.6076 ± 0.0178 & 0.6177 ± 0.0391 & 0.8002 ± 0.0180 & 0.8937 ± 0.0026 & 0.9785 ± 0.0088 & 0.9769 ± 0.0103 \\
\hline 
STAMP & 0.72M/0.72M & 0.5640 ± 0.0718 & 0.6622 ± 0.0193 & 0.8182 ± 0.0105 & 0.9172 ± 0.0100 & 0.9814 ± 0.0014 & 0.9773 ± 0.0020 \\
\hline
\end{tabular}
}
\label{tab:tuev_mumtaz}
\end{table*}

\begin{table*}[htbp]
\centering
\caption{Performance comparison of different methods on SEED-V and FACED datasets.}
\resizebox{\textwidth}{!}{%
\begin{tabular}{|l|r|ccc|ccc|}
\hline
 & & \multicolumn{3}{c|}{\textbf{SEED-V (5-Class, 117,744 Samples)}} & \multicolumn{3}{c|}{\textbf{FACED (9-Class, 10,332 Samples)}} \\
\cline{3-8}
\textbf{Methods} & \textbf{\#Params} & \textbf{Balanced Acc.} & \textbf{Cohen's Kappa} & \textbf{Weighted F1} & \textbf{Balanced Acc.} & \textbf{Cohen's Kappa} & \textbf{Weighted F1} \\
\hline
EEG Conformer & 0.55M & 0.3537 ± 0.0112 & 0.1772 ± 0.0174 & 0.3487 ± 0.0136 & 0.4559 ± 0.0125 & 0.3858 ± 0.0186 & 0.4514 ± 0.0107 \\

ST-Transformer & 3.5M & 0.3052 ± 0.0072 & 0.1083 ± 0.0121 & 0.2833 ± 0.0105 & 0.4810 ± 0.0079 & 0.4137 ± 0.0133 & 0.4795 ± 0.0096 \\
\hline
LaBraM & 5.8M & 0.3976 ± 0.0138 & 0.2386 ± 0.0209 & 0.3974 ± 0.0111 & 0.5273 ± 0.0107 & 0.4698 ± 0.0188 & 0.5288 ± 0.0102 \\

CBraMod & 15M/133M & 0.4006 ± 0.0059 & 0.2591 ± 0.0074 & 0.4101 ± 0.0065 & 0.5649 ± 0.0077 & 0.5081 ± 0.0084 & 0.5701 ± 0.0076 \\
\hline 
STAMP & 0.75M/0.76M & 0.3670 ± 0.0060 & 0.2077 ± 0.0076 & 0.3673 ± 0.0075 & 0.3606 ± 0.0061 & 0.2783 ± 0.0061 & 0.3578 ± 0.0042 \\
\hline
\end{tabular}
}
\label{tab:seedv_faced}
\end{table*}

\end{document}